\documentclass{article}

\usepackage{arxiv}

\usepackage[utf8]{inputenc} 
\usepackage[T1]{fontenc}    
\usepackage{hyperref}       
\usepackage{url}            
\usepackage{booktabs}       
\usepackage{amsfonts}       
\usepackage{nicefrac}       
\usepackage{microtype}      
\usepackage{lipsum}
\usepackage{graphicx}
\usepackage{caption} 
\usepackage{float}

\title{Tree-SNE: Hierarchical Clustering \\ and Visualization Using $t$-SNE}

\author{
  Isaac Robinson \\
  Yale University\\
  New Haven, CT 06511 \\
  \texttt{isaac.robinson@yale.edu} \\
   \And
 Emma Pierce-Hoffman \\
  Yale University\\
  New Haven, CT 06511 \\
  \texttt{emma.pierce-hoffman@yale.edu} \\
}

\begin{document}
\maketitle

\begin{abstract}
$t$-SNE and hierarchical clustering are popular methods of exploratory data analysis, particularly in biology. Building on recent advances in speeding up $t$-SNE and obtaining finer-grained structure, we combine the two to create tree-SNE, a hierarchical clustering and visualization algorithm based on stacked one-dimensional $t$-SNE embeddings. We also introduce alpha-clustering, which recommends the optimal cluster assignment, without foreknowledge of the number of clusters, based off of the cluster stability across multiple scales. We demonstrate the effectiveness of tree-SNE and alpha-clustering on images of handwritten digits, mass cytometry (CyTOF) data from blood cells, and single-cell RNA-sequencing (scRNA-seq) data from retinal cells. Furthermore, to demonstrate the validity of the visualization, we use alpha-clustering to obtain unsupervised clustering results competitive with the state of the art on several image data sets. Software is available at \url{https://github.com/isaacrob/treesne}. 
\end{abstract}

\keywords{visualization \and clustering \and hierarchical clustering \and $t$-SNE \and spectral clustering \and embedding}

\section{Introduction}

$t$-SNE (van der Maaten and Hinton 2008) is a widely-used method of visualizing high-dimensional data in low dimensions. It is motivated by minimizing the Kullback-Leibler divergence between the distributions of pairwise affinities among observations in the high-dimensional and low-dimensional spaces. Its predecessor, SNE (Hinton and Roweis 2002), uses a Gaussian kernel to transform the low-dimensional distances into affinities, while $t$-SNE uses a heavier-tailed $t$-distribution with one degree of freedom. As noted by van der Maaten and Hinton (2008), the heavier tails of the $t$-distribution compared to the Gaussian distribution help to alleviate the “crowding problem” so that distinct blobs appear in the low-dimensional embedding. 

Since $t$-SNE gives better-separated clusters than SNE, Kobak et al. (2019) used the Fourier transform (FFT)-accelerated interpolation-based $t$-SNE (FIt-SNE) approximation from Linderman et al. (2019) to implement a fast version of $t$-SNE for smaller, fractional degrees of freedom. They used a scaled $t$-distribution kernel defined as \[ k(d) = \frac{1}{(1+ d^2/\!\alpha)^\alpha}\] where the degrees of freedom are $v = 2\alpha - 1$. Since the degrees of freedom are positive ($v > 0$), $\alpha$ is also positive ($\alpha>0$). Kobak et al. found that $\alpha < 1$ (which corresponds to $v < 1$) produces $t$-SNE plots with tighter, smaller blobs, capable of capturing finer-grained structures than standard $t$-SNE. For instance, standard $t$-SNE with $\alpha = 1$ separates the images of handwritten digits from the MNIST data set by digit, whereas Kobak et al. found that two-dimensional $t$-SNE with $\alpha = 0.5$ results in further separation into blobs representing different handwriting styles of each digit (see Figure \ref{fig:MNIST-2d-t-SNE-variations}).

In a lecture at Yale University in fall 2019, Stefan Steinerberger argued that the $t$-SNE optimization problem can be interpreted as a dynamic systems problem as opposed to a probability distribution matching problem, thereby justifying using $0<\alpha<0.5$, for which the resulting kernel is not a valid probability distribution (Steinerberger 2019). He argued that this allows for an even more fine-grained look at the high dimensional data. $t$-SNE has also been shown to cluster well-separated data reliably in any embedding dimension (Linderman and Steinerberger 2019). Therefore, the current theory suggests that for well-clustered data, one-dimensional $t$-SNE can be leveraged to convey the same information as two-dimensional $t$-SNE in a more compact manner (Linderman et al. 2019). 

\begin{figure}[htp]
    \centering
    \captionsetup{width=.9\linewidth}
    \includegraphics[width=\textwidth]{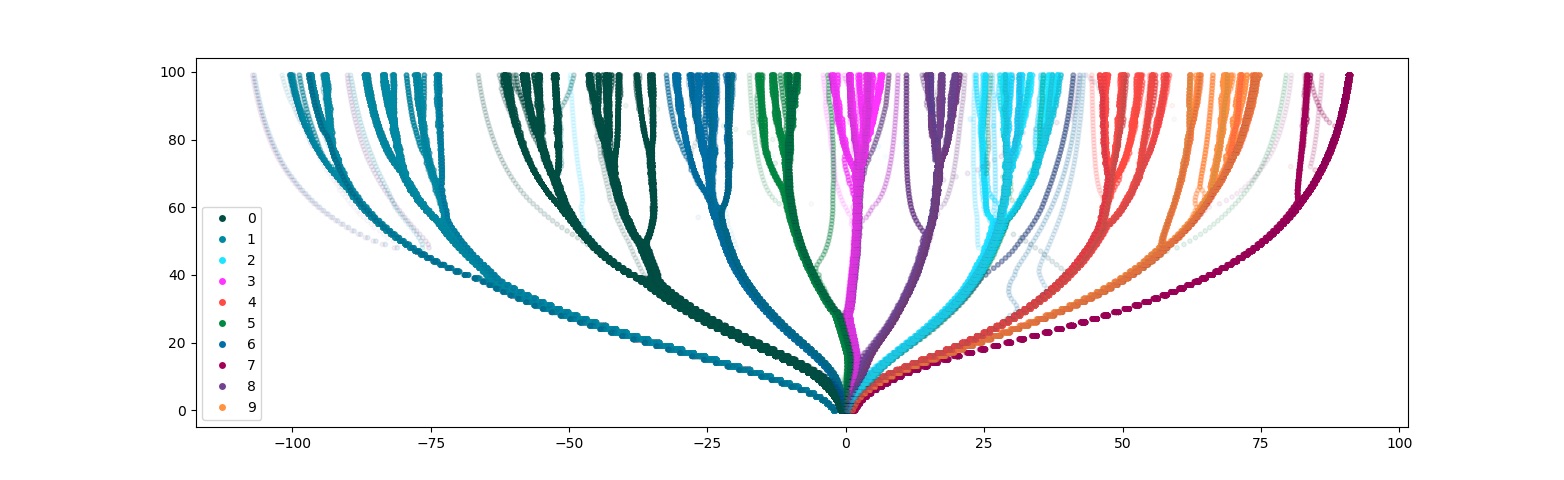}
    \includegraphics[width=\textwidth]{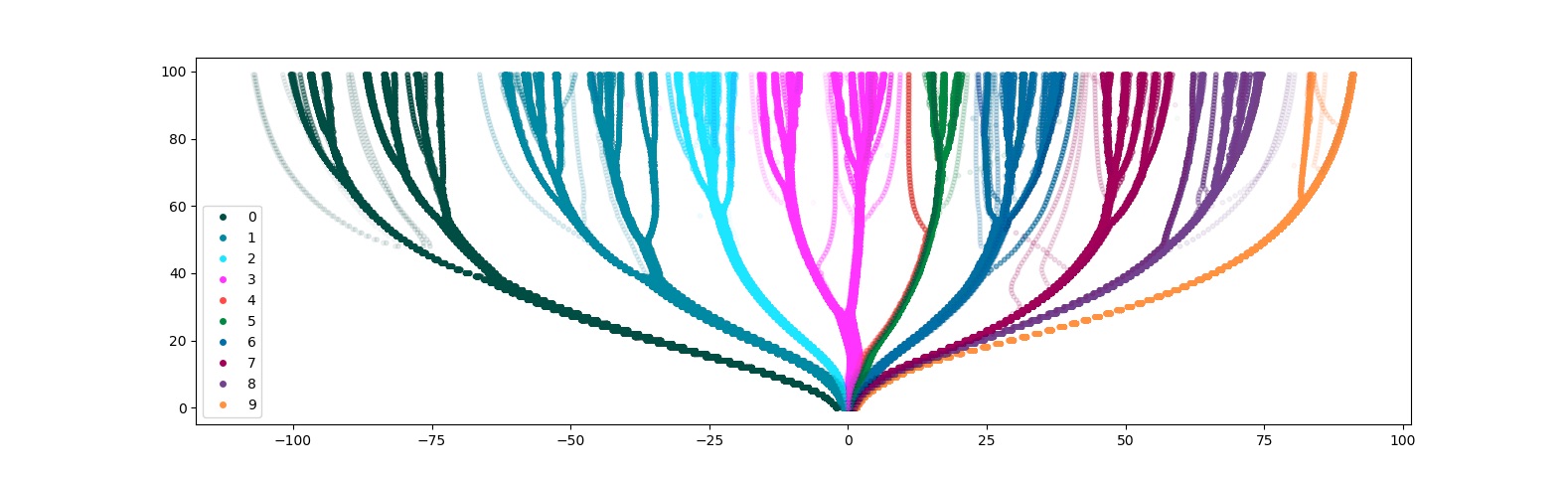}
    \caption{Tree-SNE on the USPS handwritten digits data set. Top image: the tree is colored by true digit labels. Bottom image: the tree is colored by alpha-clustering labels. Note that the alpha-clustering labels do not correspond to the digits in each cluster, which accounts for the most of the color differences between the two plots. In all tree-SNE visualizations, the y-axis shows the layer number and the x-axis is the one-dimensional $t$-SNE embedding coordinate. }
    \label{fig:USPS}
\end{figure}

Based on those results, and motivated by the popularity of $t$-SNE and hierarchical clustering individually, we present tree-SNE, a hierarchical clustering method based on one-dimensional $t$-SNE with decreasing values of $\alpha$ and perplexity at each level. Tree-SNE allows for visualization and elucidation of high-dimensional hierarchical structures by creating $t$-SNE embeddings with increasingly heavy tails to reveal increasingly fine-grained structure, and then stacking these embeddings to create a tree-like structure. We then run spectral clustering on each one-dimensional embedding, computationally determining the number of distinct clusters in the embedding. The number of clusters will increase as $\alpha$ decreases. We define the alpha-clustering of the data to be the cluster assignment that is stable across the largest range of $\alpha$ values, and then we demonstrate that alpha-clustering is competitive with the state of the art in unsupervised clustering algorithms on several data sets, including MNIST (LeCun et al. 1998) and COIL-20 (Nene et al. 1996a). Figure \ref{fig:USPS} shows example visualizations produced by tree-SNE on the USPS (Hull 1994) handwritten digits data set. 

\section{Methods}
\label{sec:headings}

In this section, we provide a description of our algorithm and motivations for various design decisions. Section 2.1 discusses tree-SNE and section 2.2 discusses alpha-clustering.

\subsection{Overview of tree-SNE}
Tree-SNE reveals hierarchical structures in high dimensional data by creating many $t$-SNE embeddings with changing perplexity and degrees of freedom of the $t$-distribution. As both perplexity and degrees of freedom decrease, smaller, more granular clusters are produced in the embedding. In this paper, we restrict ourselves to one-dimensional $t$-SNE embeddings for computational efficiency and because it makes clustering easier. In the future, two-dimensional embeddings could also be used to create 3-D tree structures that are potentially more informative. However, we will demonstrate that these one-dimensional embeddings are sufficiently informative to provide potentially valuable insight into hierarchical structures of data.

\subsubsection{Generating tree-SNE embeddings} \label{tree-sne-methods}

We begin with a standard one-dimensional $t$-SNE ($\alpha = 1$) embedding with high perplexity, by default the square root of the number of data points. A high starting perplexity increases the effective number of neighbors used by $t$-SNE (van der Maaten and Hinton 2008), which means that larger clusters will tend to form, capturing more global structures in the data (Kobak and Berens 2019). By using a large perplexity at the beginning, tree-SNE can display the entire spectrum of data organization, from global structures at the base of the tree to very fine-grained structures at the top. The default initial perplexity value of $\sqrt{N}$ was inspired by analysis of perplexity in $t$-SNE by Oskolkov (2019) and gives favorable results in practice. The exact starting value of perplexity did not appear to be important across many trials, as long as it is high, as $t$-SNE is fairly robust to small changes in perplexity (van der Maaten and Hinton 2008), and most of the interesting features of the data emerge in further embeddings from the adjustments of perplexity and $\alpha$.

Many one-dimensional $t$-SNE embeddings (typically in the range of 30 to 100) are stacked on top of each other to create the tree-SNE visualization, with the first layer at the bottom of the plot, and $\alpha$ and perplexity decreasing in each subsequent layer. To generate each sequential level, the $\alpha$ used in the previous level is multiplied by a constant factor $0<r<1$, which can be expressed as $\alpha_{n+1} = r\alpha_{n}$, where $\alpha_n$ refers to the value of $\alpha$ used in the $n$th embedding layer. With $p_n$ representing the the perplexity on level $n$, we let $p_{n+1} = p_{n}^r$. In practice, $r$ is very close to $1$. In this way, $\alpha$ approaches $0$ and perplexity approaches $1$ as the number of levels $n$ gets large, for a constant $r$. Perplexity of $1$ corresponds to entropy of $0$ in the embedding, and $\alpha$ of $0$ means our kernel function $k(x, y) = 1$ for any points $x$, $y$ in the low-dimensional embedding. Both of these parameters can be thought of as optimizing for clusters containing only a single data point each, as there is no pressure for points to form clusters under these settings. Accordingly, we observe that clusters become smaller and more numerous moving upward through the layers. 

When creating each layer, the $t$-SNE embedding is initialized with the  embedding from the previous level, rather than a standard random initialization. This is done so that each layer is a refinement of the clustering found on the previous layer, with larger clusters breaking into smaller clusters on subsequent levels of the tree. As a result, the path of a single observation or a group of observations can be traced vertically through the resulting tree visualization.  

While the FIt-SNE paper (Linderman et al. 2019) suggests that high initial exaggeration and learning rates improve the embedding quality, we use smaller learning rates and no early exaggeration after the first embedding to ensure relative stability from layer to layer. With smaller initial exaggeration and learning rates, each embedding does not change drastically from its initialization embedding, which is necessary for an interpretable tree structure to emerge. We do, however, use a constant exaggeration factor (default 12), as was done by Kobak and Berens (2019), to increase cluster separation, based on the finding by Linderman et al. (preprint 2017) that late stage exaggeration increases differentiation of clusters. 

\subsubsection{Differentiating factors of tree-SNE}

Visualizing data at multiple levels of granularity in this way can be useful both to uncover inherently hierarchical structures within the data and to explore new data, because it can be difficult to determine the optimal level of granularity on which to view or cluster unlabeled data. In standard $t$-SNE it is possible to attempt to evaluate the data across multiple scales by using different perplexity and/or degrees of freedom settings. However, the shattering of global structure is a common complaint of $t$-SNE (Kobak and Berens 2019), so it can be difficult to relate a $t$-SNE plot at one scale to another of a different scale, because the global organization of the data is not necessarily preserved. A hierarchical visualization approach such as this is necessary to enable comparison across scales. Visualizing the data with tree-SNE as opposed to $t$-SNE allows a researcher (or our alpha-clustering algorithm) to decide which scale is best for their data by showing them a more in-depth view of the hierarchical organization than is possible with standard $t$-SNE.

\subsection{Description of alpha-clustering}

Alpha-clustering extends and validates the tree-SNE algorithm by assigning cluster labels based on the $t$-SNE embeddings and then automatically selecting the best level of granularity for clustering the data, the result of which can then be compared to other unsupervised clustering algorithms to show that the structure uncovered by tree-SNE indeed captures meaningful aspects of the data. A researcher does not of course have to use alpha-clustering to use tree-SNE. However, since alpha-clustering runs on top of an expressive visualization tool, it provides a powerful opportunity for explainable unsupervised clustering which is difficult to achieve in other approaches, many of which rely on deep neural networks. It is easy to understand why alpha-clustering makes its decisions, and cluster labels can be examined on multiple levels of granularity so a researcher can choose the level that fits their needs, or the alpha-clustering algorithm will recommend an optimal level.

\begin{figure}[htp]
    \centering
    \captionsetup{width=.9\linewidth}
    \includegraphics[width=\textwidth]{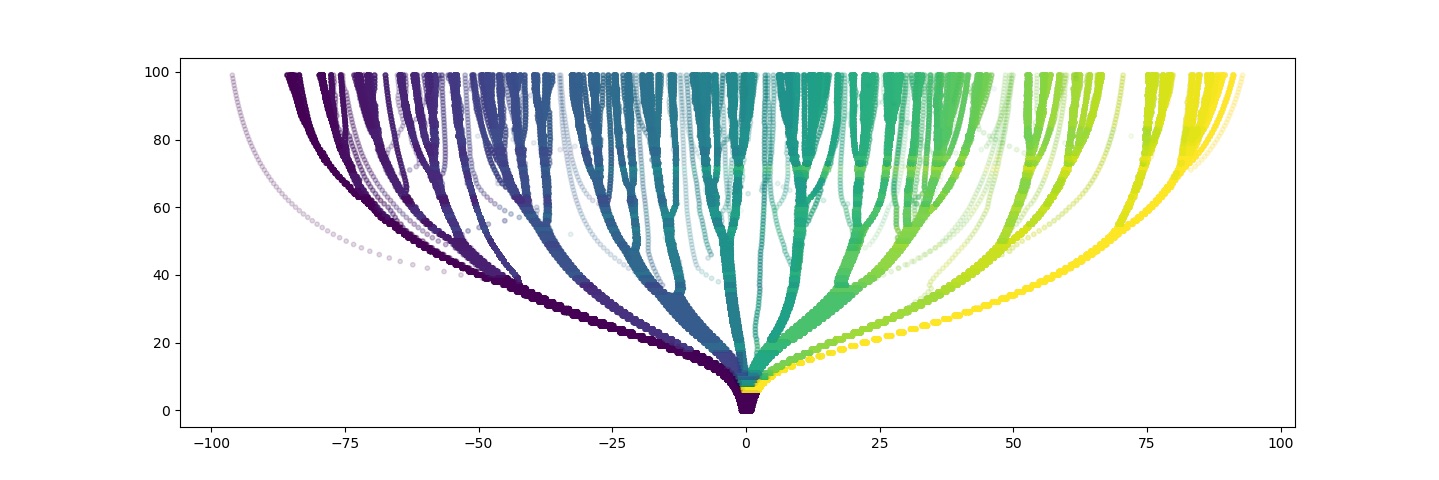}
    \includegraphics[width=\textwidth]{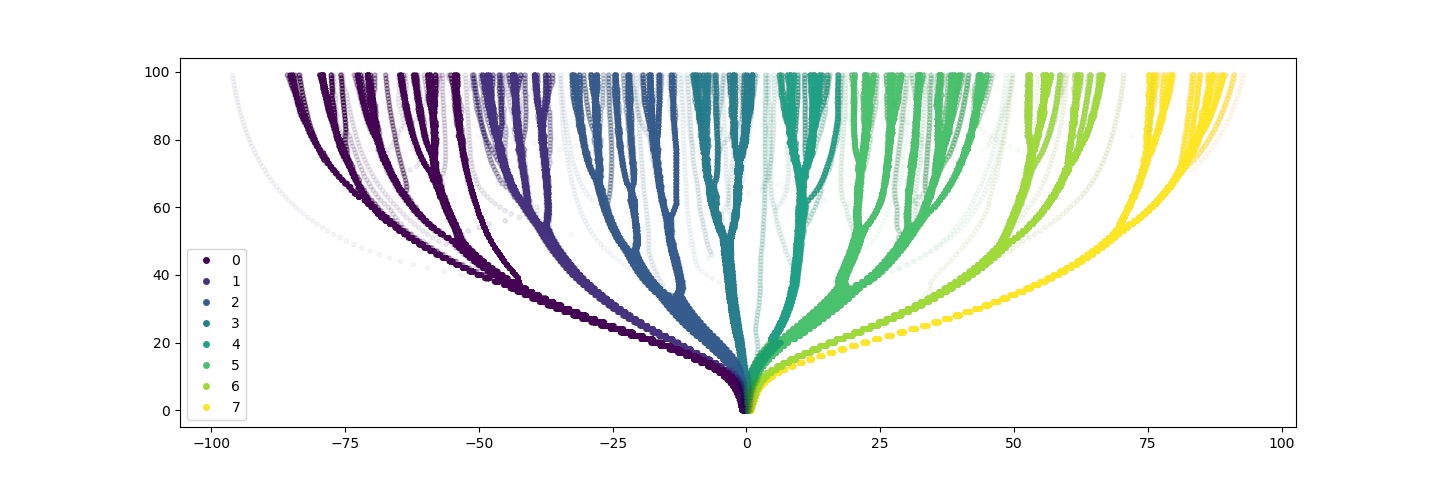}
    \caption{Tree-SNE with alpha-clustering on the MNIST handwritten digits data set. In the first visualization, each layer of the tree is colored by spectral clustering labels for that layer. In the second visualization, the tree is colored by alpha-clustering labels.}
    \label{fig:MNIST-alpha-clustering-all-best}
\end{figure}

Figure \ref{fig:MNIST-alpha-clustering-all-best} provides further insight into the alpha-clustering process. Cluster labels are assigned on each layer of the tree, and in the top image in Figure \ref{fig:MNIST-alpha-clustering-all-best}, each layer is colored by the cluster labels for that individual layer. Then, the most stable clustering out of all of the layers is defined as the alpha-clustering. In the second image in Figure \ref{fig:MNIST-alpha-clustering-all-best}, all layers of the tree are colored by the alpha-clustering labels.  

\subsubsection{Outline}

Alpha-clustering works by searching for a cluster assignment for the data that is maximally invariant across multiple levels of clustering granularity. As one-dimensional $t$-SNE tends to generate easily distinguishable clumps of data, we identify and count the clumps on each layer of the tree-SNE tree, then we define the alpha-clustering as the clustering for which the number of clumps does not change across the largest range of $\alpha$ values. 

In other words, define a clustering algorithm $c(n)$ that accepts a layer numbered $n$ and produces a clustering. If we observe that the clustering algorithm $c$ produces the same clustering on all layers between layers $n$ and $m$ (where $m > n$) such that $c(n) = c(m)$, then we call clustering $c(n)$ a stable clustering. Each level's embedding is generated by a unique value of $\alpha$, so each stable clustering has a nonzero $\alpha$ range equal to $a_m-a_n$ where $n$ and $m$ are the first and last layers that produce the stable clustering. We define the alpha-clustering to be the stable clustering for which the $\alpha$ range is the largest. Recall that the relationship between the value of $\alpha$ used to generate embeddings $n$ and $n+1$ is $\alpha_{n+1} = r\alpha_n$ for some $0<r<1$. This means that $\alpha$ varies the most in the first few layers, and as we add more layers the difference in magnitude between $\alpha_n$ and $\alpha_{n+1}$ rapidly decreases. Therefore, since alpha-clustering determines clustering stability based on the magnitude of the range of $\alpha$ values that it spans, it will tend to select for clusterings that appear earlier in the tree structure. This helps prevent alpha-clustering from over-fitting to noise in the data, as could happen if we selected a clustering that persists across multiple levels higher up the tree.

\subsubsection{Spectral clustering and determination of $k$}

Alpha-clustering uses spectral clustering on each layer of the tree. To perform spectral clustering, we first generate a graph as described in Section \ref{snn-section}. Then we compute the graph Laplacian, for which the number of zero-valued eigenvalues gives the number of connected components in the graph (von Luxburg 2006). Therefore, as one-dimensional $t$-SNE embeddings tend to give clearly distinct clusters, if we can build a graph that captures that connectivity, then the graph Laplacian will tell us the number of clusters ($k$) that are present. This allows us to avoid the need for the user to specify a number of clusters, which can be a drawback of many traditional forms of clustering when the structure of the data is unknown. It is important to note that if we were to implement a version of tree-SNE based on two-dimensional $t$-SNE embeddings, this approach would not be as successful, as it would be more difficult to build a clearly meaningfully disconnected graph on the more complex shapes that can appear in two-dimensional embeddings.

\subsubsection{Shared nearest neighbors} \label{snn-section}

For each one-dimensional embedding, we generate a graph via shared nearest neighbors (SNN) (Ertöz 2002). SNN tends to generate more disconnected graphs than k-Nearest Neighbors (k-NN), because in order for two vertices to be connected via an edge, they need to both be nearest neighbors of each other, which reduces the probability of two vertices on the edges of clusters being randomly connected. In order to reduce the probability that two distinct clusters are connected by an edge in the graph, we want to minimize the number of neighbors used; however, the number of neighbors should not be too small, because that could result in a cluster breaking into multiple disconnected components. To balance these requirements, we choose the number of neighbors to be $\beta \log{N}$, where $N$ is the number of observations, and $\beta = 2$ by default. This choice was inspired by the disconnection criterion of random graphs (Erdős and Rényi 1960) and was seen to perform well in practice. 

\subsubsection{Subsampling}

To improve the efficiency of spectral clustering on large data sets, since it is run on every layer of the tree, we only perform spectral clustering on a random subsample of 2000 observations. Then a k-nearest neighbors classifier is used to cluster the remaining observations. This greatly speeds up the implementation and does not significantly impact the performance of spectral clustering as long as we sample a large enough number of points. Using a k-nearest neighbors classifier also helps to stop individual points that appear separate from clusters from appearing to form their own clusters, by connecting them to the closest large cluster. For intuition as to why we can subsample in this way without ruining performance, see the Nyström method of kernel approximation (Williams and Seeger 2000). The Nyström method could also be used in potential future implementations of alpha-clustering to circumvent the need for a k-nearest neighbors classifier.

\subsubsection{Differentiating factors of alpha-clustering}

An important characteristic of alpha-clustering is that it does not require the user to provide $k$, the number of clusters, and instead determines the optimal level of clustering granularity from the tree-SNE embedding. In addition to the optimal alpha-clustering labels, the clustering labels at every level of the tree-SNE embedding are also returned, so the user can explore patterns on each level of granularity.

Unlike most common clustering algorithms, alpha-clustering is easily interpretable because of the accompanying visualization. The user can see the separation between clusters on each individual level, which provides more insight into the data organization than the black box of a neural network and circumvents the need for an external data visualization method to accompany other clustering algorithms like Louvain (Blondel et al. 2008) or K-Means. Not only that, but the user can see the full depth of organization of the data through the hierarchical visualization, which itself provides more flexibility to color by different features or labels than a typical dendrogram. This enables the user to explore the reasons behind specific splits in the tree and track the progression of points and clusters through the hierarchy. This helps address the increasing need for explainable machine learning methods.

\section{Demonstration on MNIST}

\subsection{Algorithm choices}

\begin{figure}[H]
    \centering
    \captionsetup{width=.9\linewidth}
    \includegraphics[width=\textwidth]{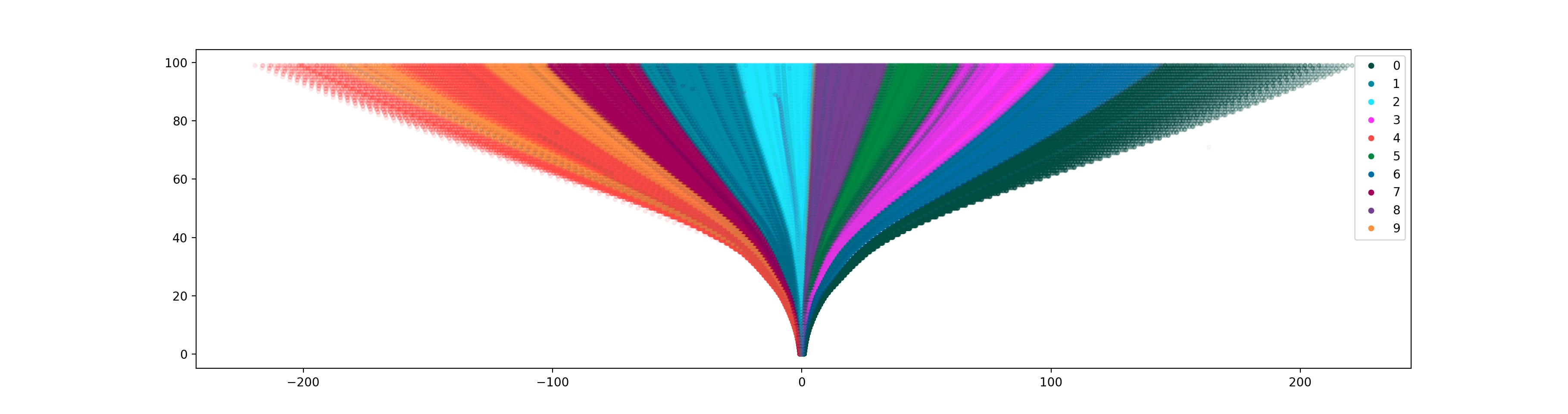}
    \includegraphics[width=\textwidth]{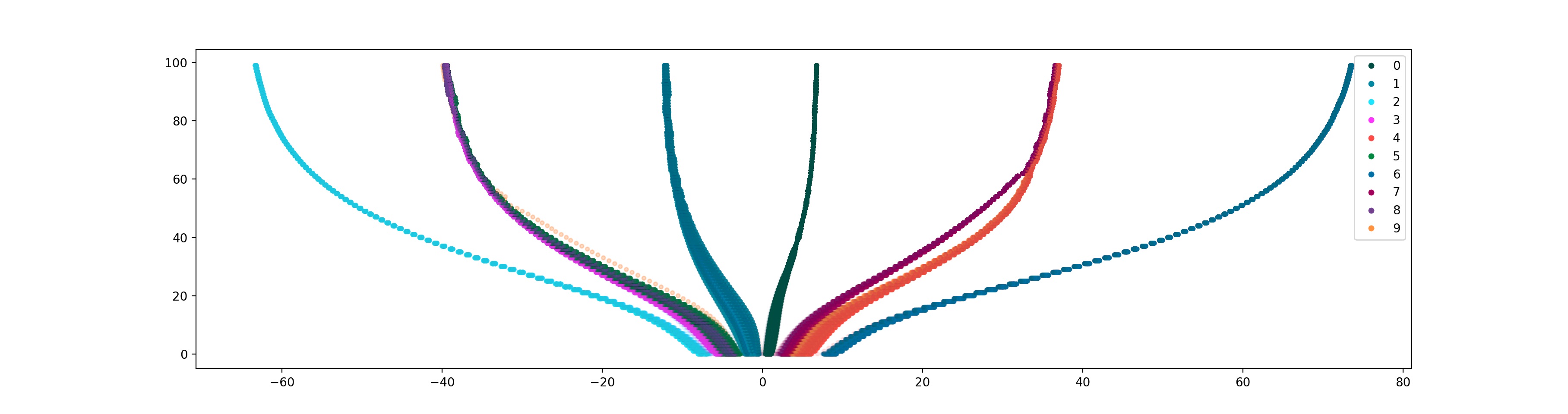}
    \includegraphics[width=\textwidth]{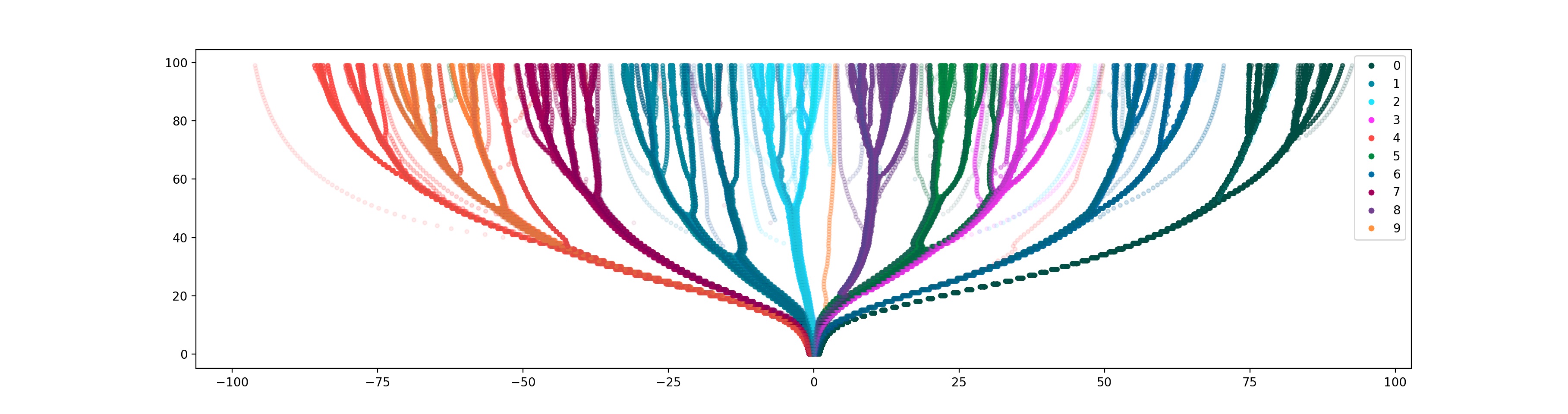}
    \caption{MNIST tree-SNE (10,000 images). In the first visualization, we do not vary degrees of freedom as we create more layers. In the second visualization, we do not vary perplexity. The last visualization shows the entire tree-SNE algorithm applied to MNIST. The data points are colored by their true labels in all three plots.}
    \label{fig:MNIST-tree-SNE-variations}
\end{figure}

To demonstrate the reasoning behind some of our algorithmic decisions, we run variations on tree-SNE with various aspects missing in Figure \ref{fig:MNIST-tree-SNE-variations}. 

In the top visualization in Figure \ref{fig:MNIST-tree-SNE-variations}, we run tree-SNE without decreasing the degrees of freedom at each level, meaning we only decrease perplexity. Notice how the data is organized and for the most part captures the main structure of MNIST; the visualization is split into different colors showing the grouping of different MNIST digits, and except for fours and nines they are cleanly divided. However, there is no cluster separation, and the visualization does not divide the structure into any substructures. 

\begin{figure}[htp]
    \centering
    \captionsetup{width=.9\linewidth}
    \includegraphics[width=0.9\textwidth]{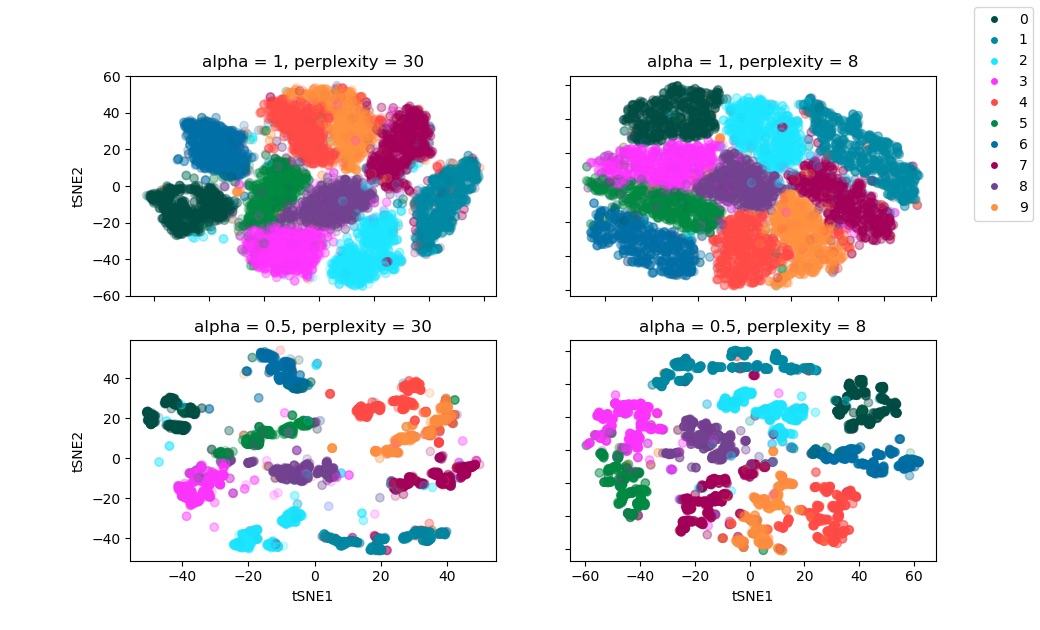}
    \caption{Two-dimensional t-SNE on 10,000 MNIST digits. The top row has $\alpha$ = 1 and the bottom row has $\alpha = 0.5$. The left column has perplexity = 30, while the right column has perplexity = 8. The plots are colored by truth labels for the digits.}
    \label{fig:MNIST-2d-t-SNE-variations}
\end{figure}

The middle visualization in Figure \ref{fig:MNIST-tree-SNE-variations} does not decrease perplexity, and only decreases the degrees of freedom ($\alpha$). The clusters become tighter as $\alpha$ decreases, but they do not split into smaller clusters. We observe in Figure \ref{fig:MNIST-2d-t-SNE-variations} that in two-dimensional $t$-SNE, decreasing $\alpha$ is sufficient to reveal cluster substructure, and decreasing perplexity in tandem is not necessary, as was previously shown in Kobak et al. (2019). However, we see in the middle plot of Figure \ref{fig:MNIST-tree-SNE-variations} that decreasing $\alpha$ alone is not enough to produce finer-grained clusters in one-dimensional $t$-SNE embeddings. 

The final visualization in Figure \ref{fig:MNIST-tree-SNE-variations} is our technique, tree-SNE, which decreases $\alpha$ and perplexity in tandem through the layers. Notice how clusters on earlier levels break into smaller subclusters on later levels, unlike the two other visualizations. We observe that decreasing both $\alpha$ and perplexity appears to be necessary to reveal cluster substructures in one-dimensional $t$-SNE, and tree-SNE therefore takes this approach,  balancing the spreading and contracting forces of decreasing perplexity and the degrees of freedom to build a tree structure containing meaningful hierarchy.

\subsection{Showing hierarchy} \label{sec:hierarchy}

\begin{figure}[H]
    \centering
    \captionsetup{width=.9\linewidth}
    \includegraphics[width=\textwidth]{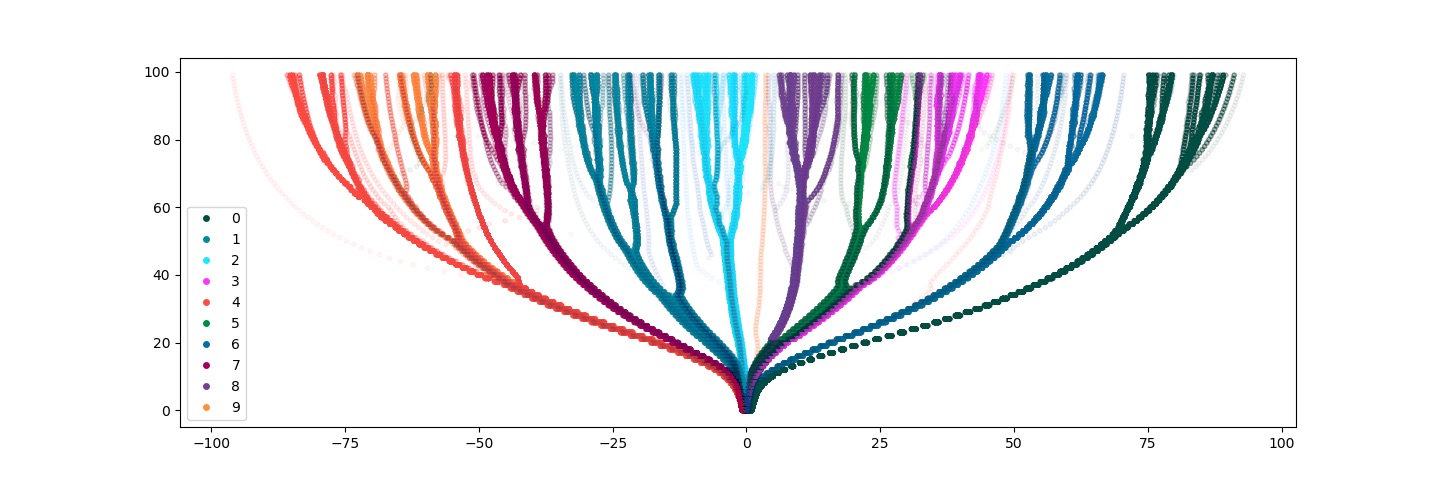}
    \includegraphics[width=\textwidth]{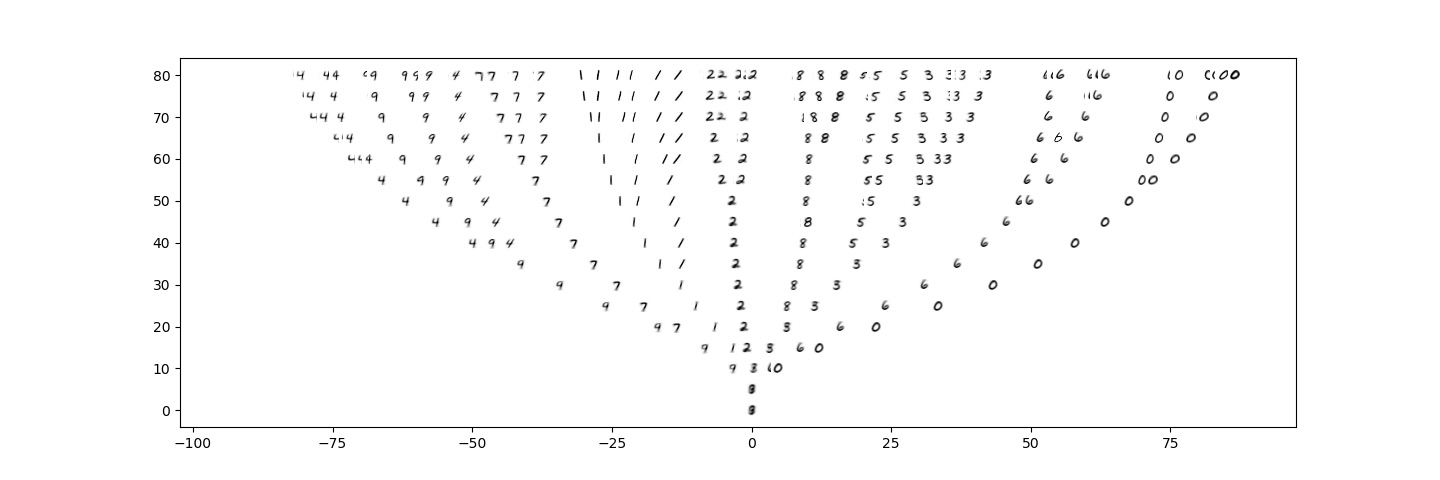}
    \includegraphics[width=0.9\textwidth]{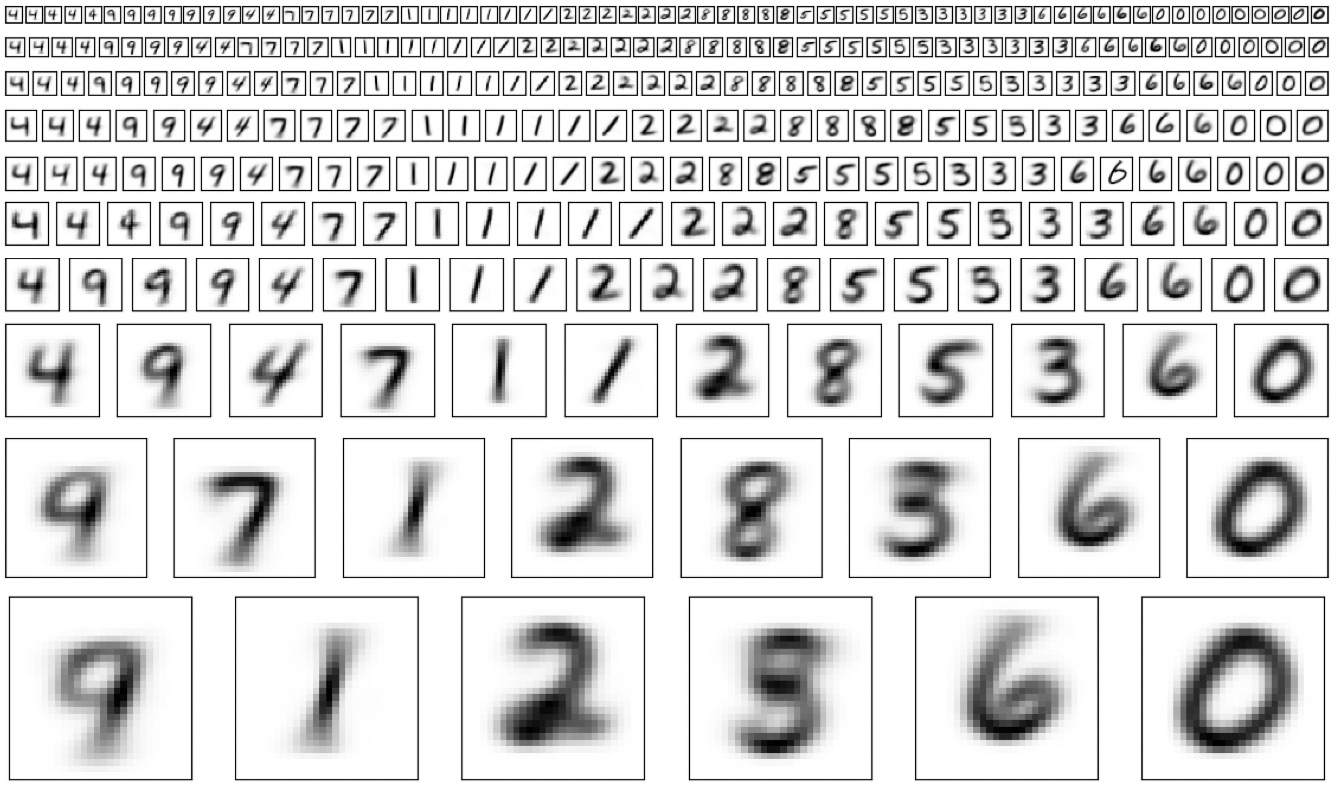}
    \caption{MNIST (10,000 samples) tree-SNE with averaged images of digits within each cluster. The top plot is a tree-SNE visualization of MNIST colored by the true digit labels. The middle plot shows the within-cluster average digits superimposed on the tree structure. The bottom visualization shows within-cluster average digits for a sample of layers. Clusters determined via alpha-clustering on each layer.}
    \label{fig:MNIST-hierarchy-example}
\end{figure}

In order to illustrate how tree-SNE displays hierarchical organization of data, we visualized the MNIST handwritten digits data set, then we examined the averaged digit images for each cluster on 17 of the levels (every fifth level between zero and 80, inclusive). The results are shown in two different formats in Figure \ref{fig:MNIST-hierarchy-example}. 

The top plot in Figure \ref{fig:MNIST-hierarchy-example} shows the tree-SNE visualization of MNIST data that was used for this analysis of hierarchical organization. It was run with a different random seed than the visualization in Figure \ref{fig:MNIST-tree-SNE-variations}, but it is highly similar in appearance. This plot is colored by true digit labels, and we observe that the digits separate cleanly for the most part by level 21, the level chosen by alpha-clustering, with two notable exceptions: four is grouped with nine, and three is grouped with five. In both cases, these digits split farther up the tree. It makes sense that these pairs of numbers remain un-separated longer than other digits, considering their similar appearances. 

The middle plot in Figure \ref{fig:MNIST-hierarchy-example} shows within-cluster average digits for each cluster identified by spectral clustering on each of 17 levels of the tree-SNE visualization from the top image. Each averaged image is plotted in the location corresponding to the middle of the cluster from which it was generated, so that the images can be viewed in the tree-SNE tree structure. We observe that the clusters are refined moving up the tree, as $\alpha$ and perplexity decrease. Clusters representing multiple digits separate by digit, and clusters representing a single digit split into smaller clusters representing different handwriting variations for that digit. For instance, ones split into vertical and tilted versions around level 35, and twos split into bubbly and straight sub-groups at level 55. There is a clear hierarchy of digit types, with distinctions becoming finer-grained at the top of the plot. From this visualization, we see that tree-SNE's branching structure conveys meaningful information about the hierarchical organization of the data.

Some overlap between images was unavoidable while preserving the tree structure, so the bottom visualization in the same figure (Figure \ref{fig:MNIST-hierarchy-example}) is provided for larger views of the within-cluster average digits. Within-cluster average digits are shown for ten of the 17 layers shown in the middle plot, stacked vertically in the same order as they appear in the tree-SNE visualization. The second row from the bottom represents the stable clustering identified by alpha-clustering, in which only fours and nines and threes and fives have yet to separate cleanly. We draw attention to the top row, where we observe that the rightmost cluster of fours that is last to split from nines is triangular in shape, so in fact is more similar in appearance to nines than to the other fours. Based on these visualizations of within-cluster averaged digits, we see that tree-SNE produces meaningful clusters on each embedding, with hierarchical organization from layer to layer.

\subsection{Alpha-clustering}

\begin{figure}[htp]
    \centering
    \captionsetup{width=.9\linewidth}
    \includegraphics[width=\textwidth]{figures/mnist_true_labels_widest.jpg}
    \includegraphics[width=\textwidth]{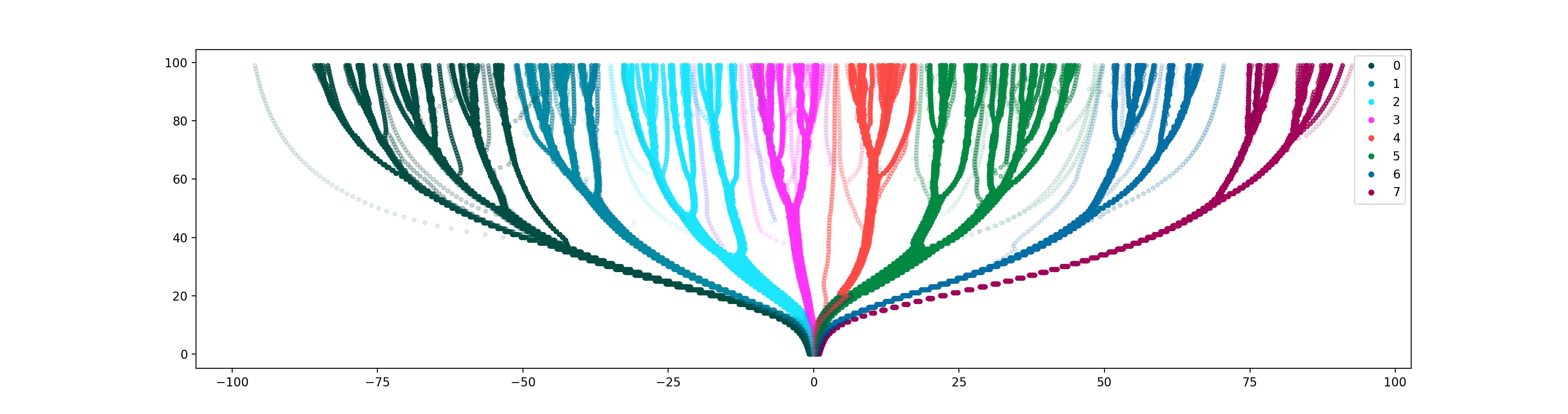}
    \caption{MNIST tree-SNE embeddings. The top image is colored by true digit labels, and the bottom image is colored by alpha-clustering labels. As with Figure \ref{fig:USPS}, note that the alpha-clustering labels do not correspond to the digits in each cluster, which accounts for the most of the color differences between the two plots.}
    \label{fig:MNIST-clusters}
\end{figure}

We apply alpha-clustering to MNIST. Recall that alpha-clustering determines the level with the clustering that best fits the overall structure of the data. Figure \ref{fig:MNIST-clusters} shows the tree-SNE embedding of MNIST from Figure \ref{fig:MNIST-tree-SNE-variations} in the top image, and the same embedding with data points on all layers colored by alpha-clustering labels in the bottom plot. As discussed in Section \ref{sec:hierarchy}, alpha-clustering correctly groups six of the ten MNIST classes, and combines the remaining four into two different clusters of two each. It combines four and nine, and three and five. Looking closer at the tree structure in the top plot of Figure \ref{fig:MNIST-clusters}, we see that the split that separates three from five happens at roughly the same level that one splits into two different sub-classes. Since alpha-clustering selects the optimal cluster assignment from a single layer, choosing a clustering level that separates three and five would mean also subdividing the ones into multiple clusters. 

In order to determine the quality of a clustering assignment, we consider the normalized mutual information (NMI) (Strehl and Ghosh 2002) of a clustering. The NMI is defined as \[\frac{I(X; Y)}{\sqrt{H(X)H(Y)}}\] where $X$ and $Y$ are two different cluster assignments for the data, $I$ is the mutual information, and $H$ is the entropy. Values of NMI fall between $0$ and $1$, with $1$ representing a complete match between $X$ and $Y$. We can assess the quality of alpha-clustering by computing its NMI relative to the ground truth labels: $X$ = alpha-clustering labels and $Y$ = true labels. On MNIST (10,000 samples), alpha-clustering has an NMI score of 0.84. Importantly, there are is no clustering assignment on any individual layer with a higher NMI. This shows that alpha-clustering is selecting the best possible clustering from all levels of the tree. 

\section{Experiments}

We demonstrate tree-SNE on several popular clustering data sets and two biological data sets. The purpose of running tree-SNE on the clustering data sets is to show that it can generate a competitive and interpretable clustering, which means that one-dimensional $t$-SNE embeddings created in this way are indeed able to express meaningful organization of high-dimensional data. Applying tree-SNE to biological data sets allows us to demonstrate its performance on diverse types of data and compare it to standard visualization techniques in biology. 

\subsection{Experimental Setup}

Tree-SNE has three major parameters: the ratio $r$, the initial perplexity, and the number of layers. For the following experiments, we set the starting perplexity to be the square root of the number of observations in the data set ($\sqrt{N}$), as described in Section \ref{tree-sne-methods}. Tree-SNE embeddings tend to be exhibit highly  branched structures corresponding to very local structures in the data for $\alpha<0.01$, so by default we set the lower bound on $\alpha$ to 0.01. Given this fixed lower bound on $\alpha$, we can solve for the ratio $r$ required to decrease $\alpha$ from 1 to 0.01 in a given number of layers $n$: \[ r = e^{\frac{\ln{0.01}}{ n}}\]
Using these sensible defaults, $r$ and perplexity are automatically determined for a given data set and a desired number of layers for the tree. For testing alpha-clustering without visualization, we use 30 layers, and for visualization, we use 100 layers. Using fewer layers has minimal effect on the alpha-clustering and is more computationally efficient for repeated experimentation, while a larger number of layers is ideal for visualization because it produces a higher-resolution plot. For example, for 30 layers, $r$ is $0.858$, and for 100 layers, $r$ is $0.955$. Alpha-clustering has one major parameter, $\beta$, where $\beta \log{n}$ is the number of nearest neighbors used to generated the shared nearest neighbors graph for spectral clustering. We report our results with the default of $\beta = 2$, unless otherwise noted.

\subsection{Results of alpha-clustering}

We apply alpha-clustering to four standard clustering data sets and compare the NMI to that of several state of the art clustering methods. The results are shown in Table \ref{tab:alpha-clustering}, which is based on Table 4 from Gultepe and Makrehchi (2008). Alpha-clustering achieves a competitive NMI on all of the data sets.

\begin{table}[H]
 \caption{Performance of different clustering algorithms}
  \centering
  \begin{tabular}{lllll}
    \toprule
    Method     & COIL-20     & COIL-100     & USPS & MNIST \\
    \midrule
    \multicolumn{1}{c}{Deep Learning} \\
    \cmidrule(r){1-1}
    JULE* (2016) & \bf 1.000  & \bf 0.985    & \bf 0.913 & 0.913     \\
    DBC* (2017)     & 0.895  & 0.905 & 0.724 & \bf 0.917  \\
    IEC* (2016)    & - & 0.787 & 0.641 & 0.542      \\
    DEPICT* (2017)    & - & - & \bf 0.927 & \bf 0.917     \\
    AEC* (2013)  & - & - & 0.651 & 0.669 \\
    NMF-D* (2014) & 0.692 & 0.719 & 0.287 & 0.152 \\
    TSC-D* (2016) & 0.928 & - & - & 0.651 \\
    DCEC* (2017) & - & - & 0.826 & 0.885 \\
    SpectralNet* (2018) & - & - & - & \bf 0.924 \\
    \cmidrule(r){1-1}
    \multicolumn{1}{c}{Non-deep learning} \\
    \cmidrule(r){1-1}
    ICA BSS* (2018)     & \bf 0.965 & \bf 0.962 & 0.868 & 0.824      \\
    AC-PIC* (2013)     & 0.855 & 0.840 & 0.840 & 0.017      \\
    K-means* & 0.774 & 0.775 & 0.613 & 0.490 \\ 
    DBSCAN & 0.892 & 0.705 & 0.299 & - \\
    \cmidrule(r){1-1}
    Ours        & \bf 0.958 & \bf 0.926** & \bf 0.885 & 0.864 \\ 
    \bottomrule
  \end{tabular}
  \captionsetup{width=.9\linewidth}
  \caption*{Table 1. The NMI scores of various clustering algorithms on different data sets, with top three in bold. For DBSCAN, $\varepsilon$ was tuned until the correct number of clusters was obtained. We performed DBSCAN, K-Means, and tree-SNE benchmarking; the rest of the values are reported from the original papers. * means the clustering algorithm requires knowing the number of clusters beforehand. ** this was run with $\beta=1.3$, whereas the rest were run with the default $\beta=2$. The methods in the table are: Joint Unsupervised Learning (JULE, Yang et al. 2016), Discriminatively Boosted Clustering (DBC, Li et al. 2018), Infinite Ensemble Clustering (IEC, Liu et al. 2016), Autoencoder-based Clustering (AEC, Song et al. 2013), NMF with Deep learning model (NMF-D, Trigeorgis et al. 2014), Task-specific Deep Architecture for Clustering (TSC-D, Wang et al. 2016), Deep Convolutional Embedded Clustering (DCEC, Guo et al. 2017), SpectralNet (Shaham et al. 2018), Independent Component Analysis Blind Source Separation (ICA BSS, Gultepe and Makrehchi 2018), Agglomerative Clustering via Path Integral (AC-PIC, Zhang et al. 2013), K-Means, Density-Based Spatial Clustering of Applications with Noise (DBSCAN, Ester et al. 1996). Benchmarking data sets: COIL-20 (Nene et al. 1996a), COIL-100 (Nene et al. 1996b), USPS (Hull 1994), MNIST 60,000 samples (LeCun et al. 1998).}
  \label{tab:alpha-clustering}
\end{table}

Note that while most of the other methods featured in Table \ref{tab:alpha-clustering} require specifying a number of clusters, alpha-clustering determines the best number of clusters from the data.
In biomedical applications, the optimal number of clusters is often not known {\em a priori}, so algorithms such as tree-SNE that do not require the number of clusters as input may be particularly useful.

If the number of clusters is known beforehand, the $\beta$ parameter can be tuned as needed such that alpha-clustering yields closer to the correct number of clusters. This is particularly useful when the data has a large number of clusters, as some clusters may be artificially merged in the SNN graph, and a lower $\beta$ value will decrease the number of neighbors used to construct the graph and reduce this merging behavior. We find that decreasing $\beta$ from our default of $2$ to $1.3$ increases the performance of of alpha-clustering on COIL-100 from $0.899$ to $0.926$ and increases the number of clusters found from 76 to 101. Note that changing $\beta$ to give the correct number of clusters does not always increase the NMI, as the clusters that split apart with lower $\beta$ do not necessarily correlate with true label clusters. For example, tuning $\beta$ on COIL-20 to give the correct number of clusters results in a decreased NMI of 0.951.

It should be noted that all of the benchmarking data sets used are image data sets: the two Columbia Object Image Library data sets, COIL-20 (Nene et al. 1996a) and COIL-100 (Nene et al. 1996b); as well as two handwritten digit data sets, USPS (Hull 1994) and MNIST (LeCun et al. 1998). These data sets were chosen due to readily available comparison metrics from other methods. However, convolutional neural network-based approaches have an advantage on these data sets because they are designed to process image data.

\subsection{Results of tree-SNE}

We demonstrate running tree-SNE on several different data sets, including mass cytometry (CyTOF) data from bone marrow cells (Bendall, et al. 2011) and single-cell RNA-sequencing (scRNA-seq) data from retinal cells (Shekhar et al. 2016). We compare tree-SNE with $t$-SNE (van der Maaten and Hinton 2008) and PHATE (Moon et al. 2019), standard approaches in biomedical data visualization.

\subsubsection{CyTOF}

\begin{figure}[htp]
    \centering
    \captionsetup{width=.9\linewidth}
    \includegraphics[width=\textwidth]{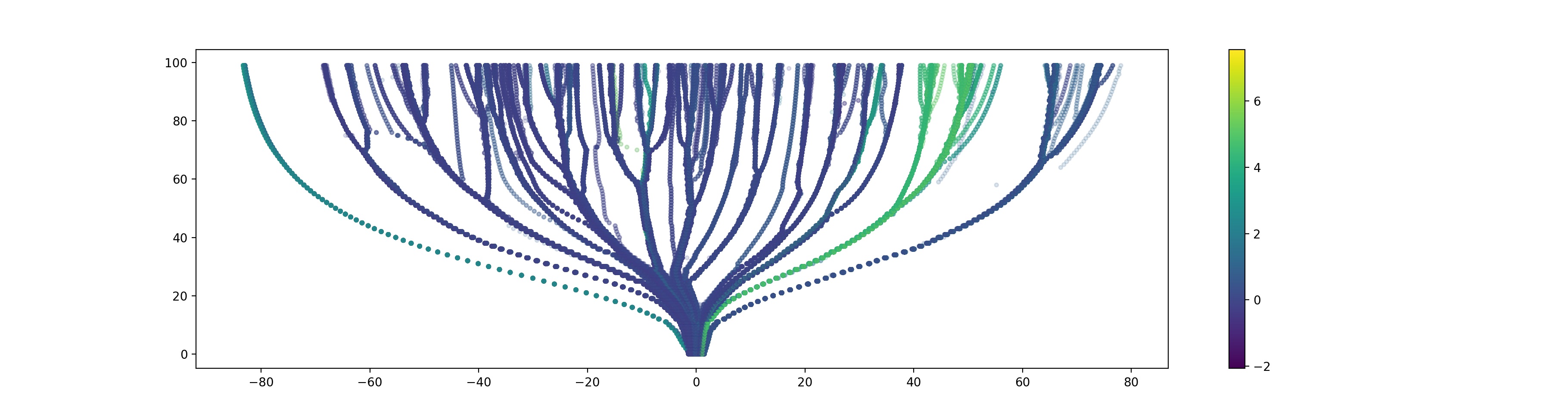}
    \includegraphics[width=\textwidth]{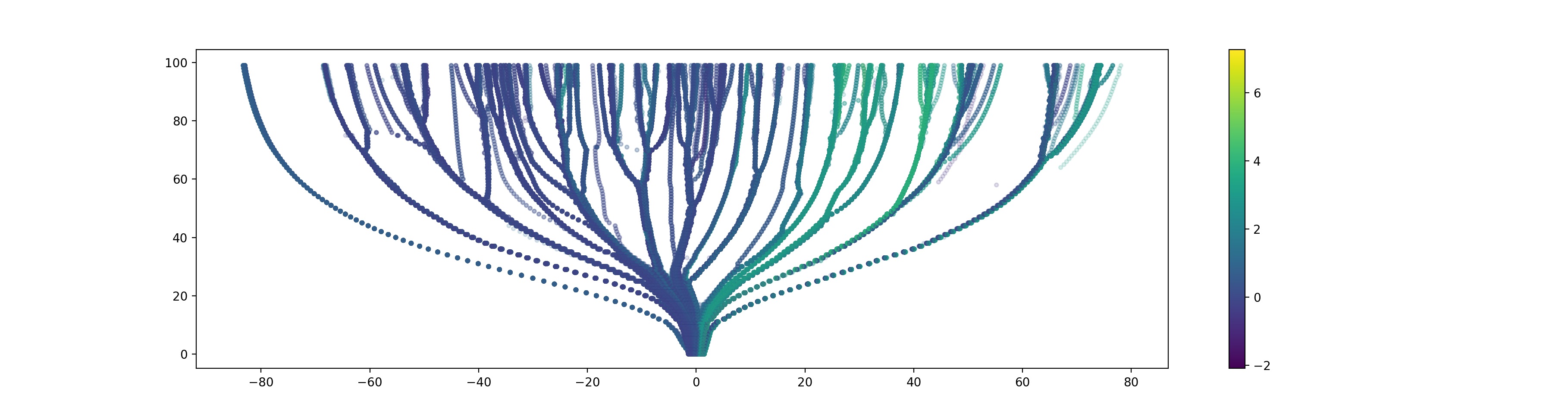}
    \caption{CyTOF tree-SNE embeddings. The top embedding is labeled by the level of CD8 in the sample, and the bottom is labeled by CD45ra expression.}
    \label{fig:CyTOF-tree-SNE}
\end{figure}

Figure \ref{fig:CyTOF-tree-SNE} shows a tree-SNE embedding of bone marrow single-cell mass cytometry (CyTOF) data (Bendall et al. 2011), colored by the expression levels of specific markers in each cell. The top visualization is colored by intensity of CD8, which is a marker for cytotoxic T cells, and the bottom is colored by intensity of CD45ra, which is a marker for naive T cells (Golubovskaya and Wu 2016). Notice how in the top visualization most of the cytotoxic T cells are in the same bright arm, which subdivides at higher embedding levels. This indicates that tree-SNE separates cytotoxic T cells on lower levels of the embedding, and then splits them into subgroups higher up as $\alpha$ decreases. Importantly, the CD8 branch splits in two about halfway up the plot. The CD45ra visualization (bottom of Figure \ref{fig:CyTOF-tree-SNE}) reveals that the split in the cytotoxic T cell line is mediated by the presence of CD45ra: CD45ra is present in one of the branches but not the other. This indicates that tree-SNE is splitting cytotoxic T cells ($\textrm{CD8}^+$) into naive ($\textrm{CD45ra}^+$) and mature cells ($\textrm{CD45ra}^-$). In this way, tree-SNE very clearly reveals the hierarchical organization of the CyTOF data and facilitates identification of the features mediating the branching. 

\begin{figure}[htp]
    \centering
    \captionsetup{width=.9\linewidth}
    \begin{minipage}[b]{0.4\textwidth}
    \includegraphics[width=200pt]{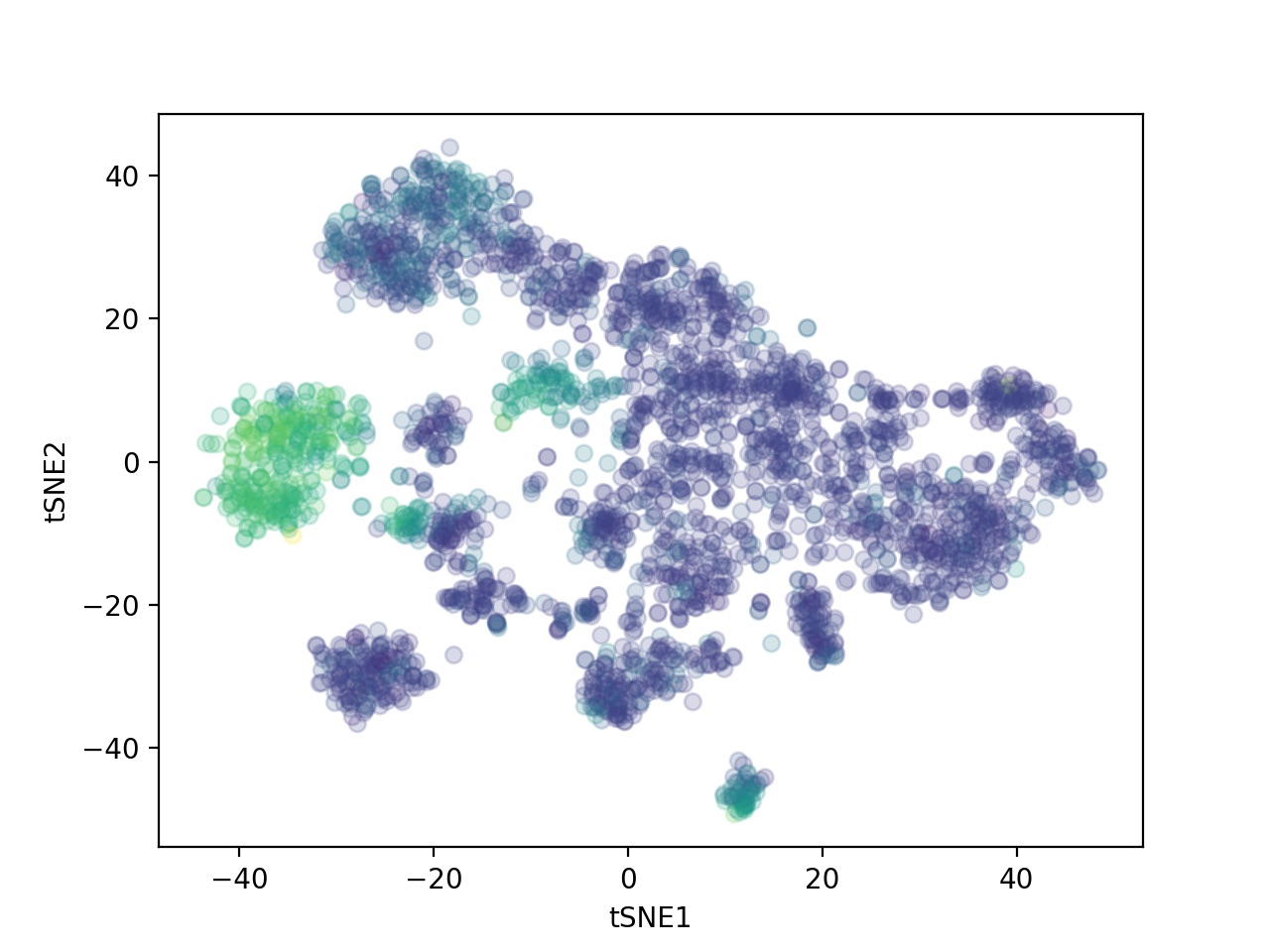}
    \end{minipage}
    \begin{minipage}[b]{0.4\textwidth}
    \includegraphics[width=200pt]{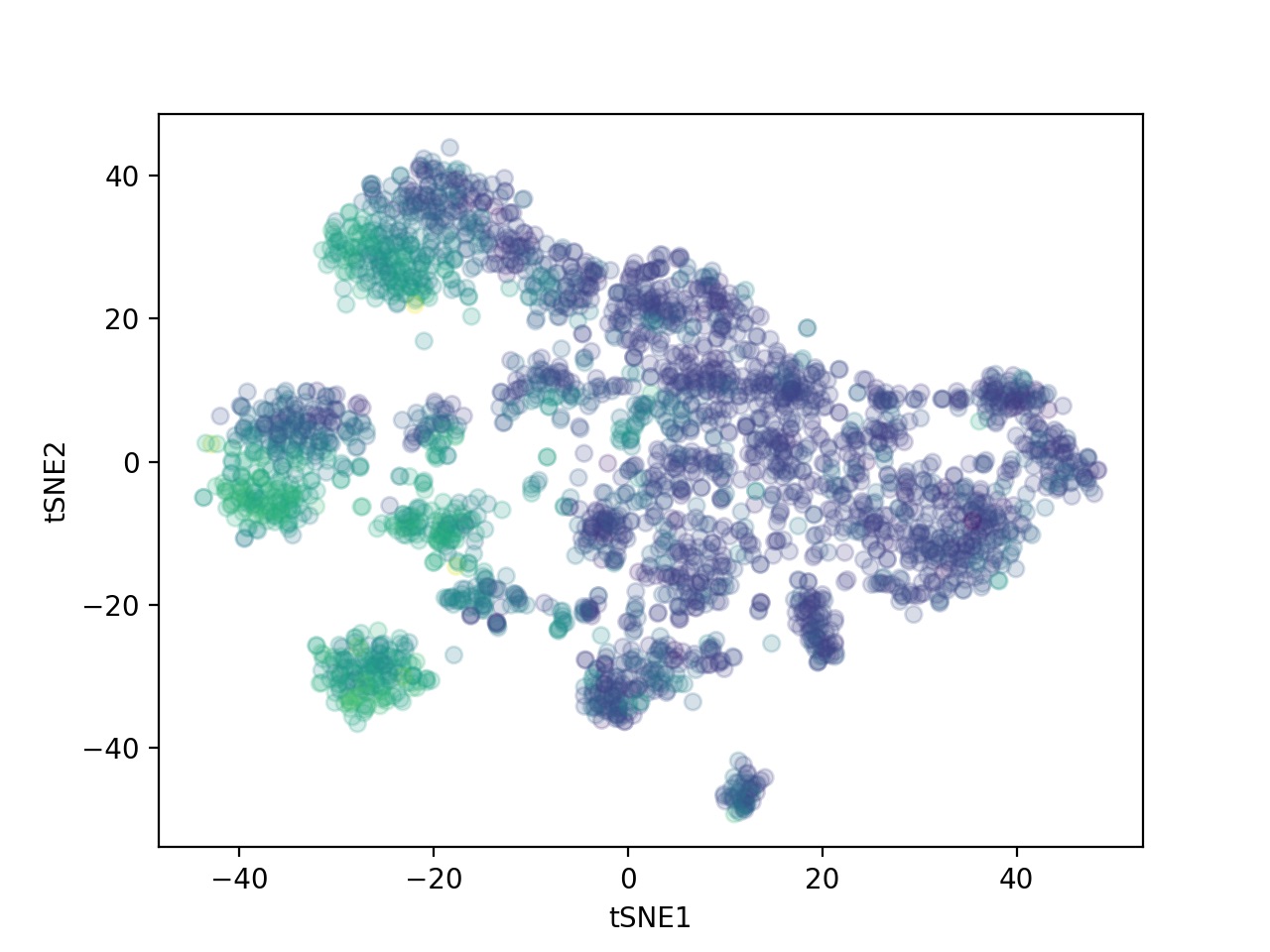}
    \end{minipage}
    \caption{CyTOF $t$-SNE embeddings. The left is colored by CD8, and the right is colored by CD45ra.}
    \label{fig:CyTOF-$t$-SNE}
\end{figure}

\begin{figure}[htp]
    \centering
    \captionsetup{width=.9\linewidth}
    \begin{minipage}[b]{0.4\textwidth}
    \includegraphics[width=200pt]{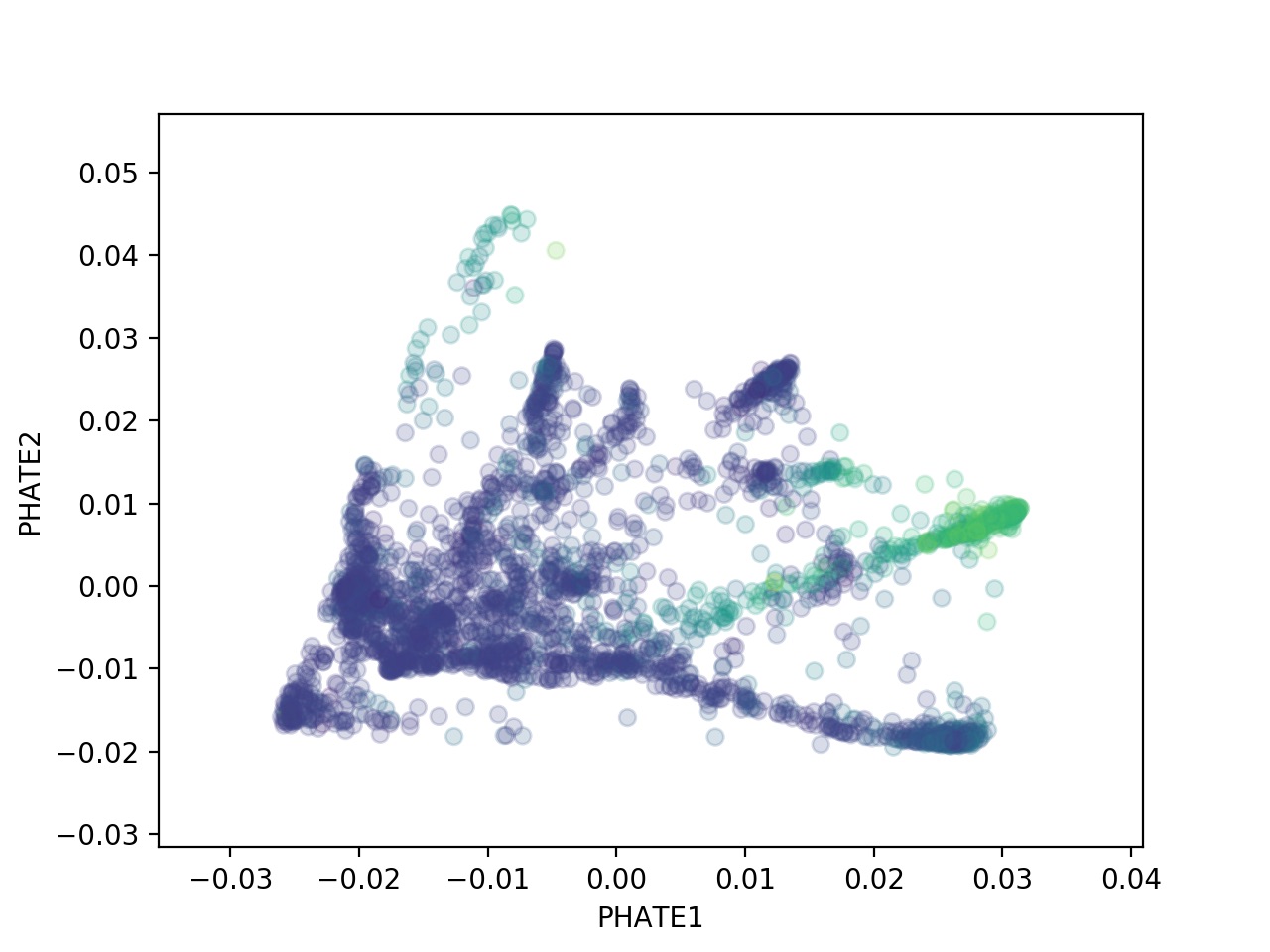}
    \end{minipage}
    \begin{minipage}[b]{0.4\textwidth}
    \includegraphics[width=200pt]{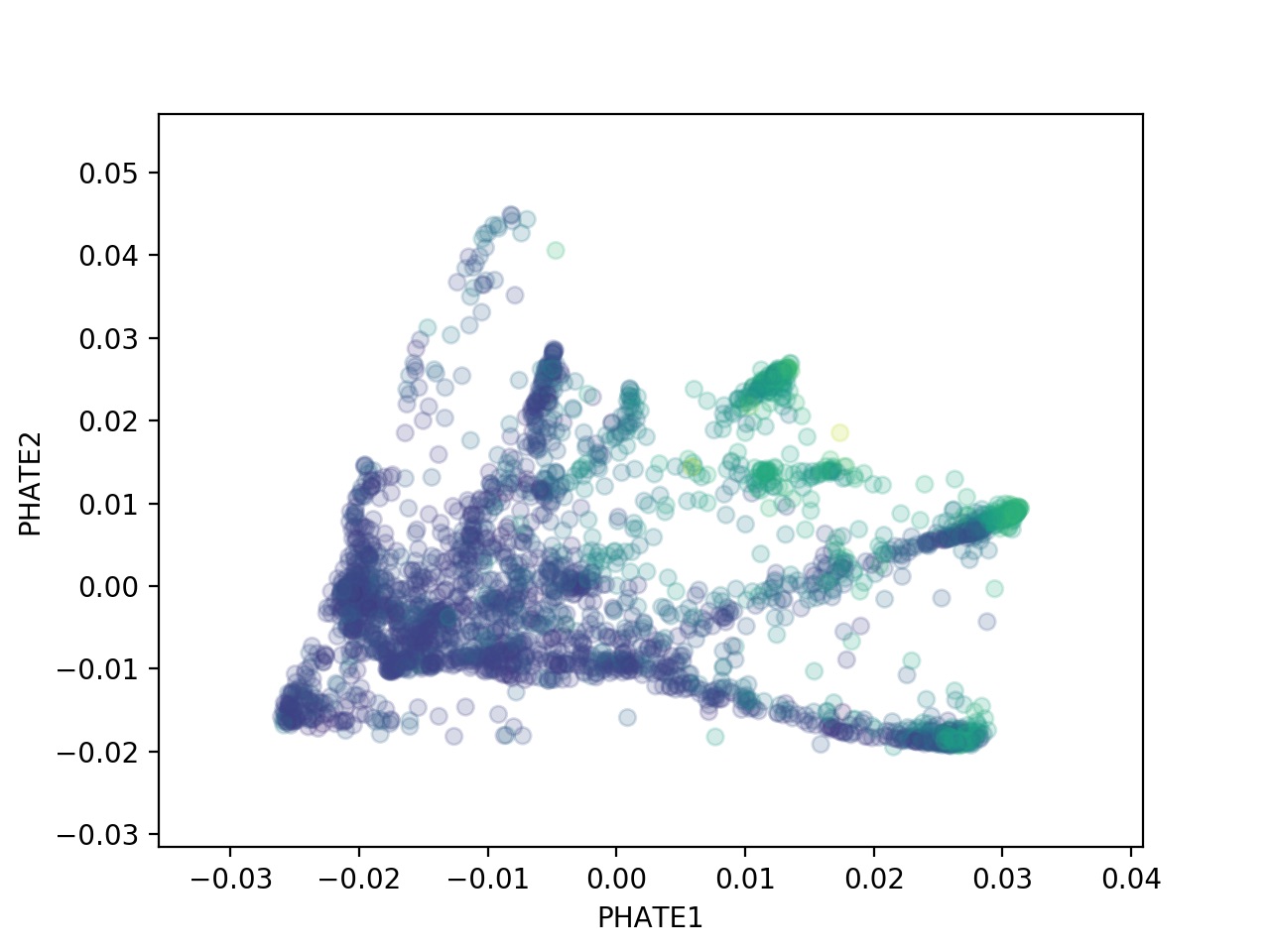}
    \end{minipage}
    \caption{CyTOF PHATE embeddings. The left is colored by CD8, and the right is colored by CD45ra.}
    \label{fig:CyTOF-PHATE}
\end{figure}

For comparison with established visualization methods for biological data, we show the same data embedded via $t$-SNE (van der Maaten and Hinton 2008) in Figure \ref{fig:CyTOF-$t$-SNE} and via PHATE (Moon et al. 2019) in Figure \ref{fig:CyTOF-PHATE}. 

\subsubsection{scRNA-seq} \label{sec:shekhar}

We demonstrate tree-SNE on single-cell RNA-sequencing (scRNA-seq) data from mouse retinal cells and compare the results with the findings from the original paper by Shekhar et al. (2016). Shekhar et al. used Louvain community detection (Blondel et al. 2008) to detect clusters, two-dimensional $t$-SNE to visualize the clusters, and hierarchical clustering with Euclidean distances and average linkage to produce a hierarchical depiction of the relationships between different mouse retinal bipolar cells (BCs). We run tree-SNE on the same data and label it according to the labels assigned by Shekhar et al. from clustering and morphological analysis. Some pre-processing, mediated by alpha-clustering, was performed to roughly isolate the cone bipolar cells, which encompass the cell subtypes evaluated by hierarchical clustering by Shekhar et al. The full process is described in Appendix A (Section \ref{sec:shekhar-appendix}). Figure \ref{fig:scRNA-seq-tree-SNE} shows the tree-SNE embedding of the data subset. 

Although the tree-SNE hierarchy and the dendrogram created by Shekhar et al. do not match entirely, there are strong similarities between the two. Notice how BC1A and BC1B split apart at a high level of the tree, and how they are more closely related to BC2, BC3A, BC3B, and BC4 than to the BC5 subtypes or BC7. Observe also how tree-SNE show the same organization of the four subtypes of BC5 as Shekhar et al. discovered: BC5A and BC5D are closely related, and BC5B and BC5C are closely related. Tree-SNE also shows further branching of some of the cell types, and these subdivisions may warrant further examination.

\begin{figure}[htp]
    \centering
    \captionsetup{width=.9\linewidth}
    \includegraphics[width=\textwidth]{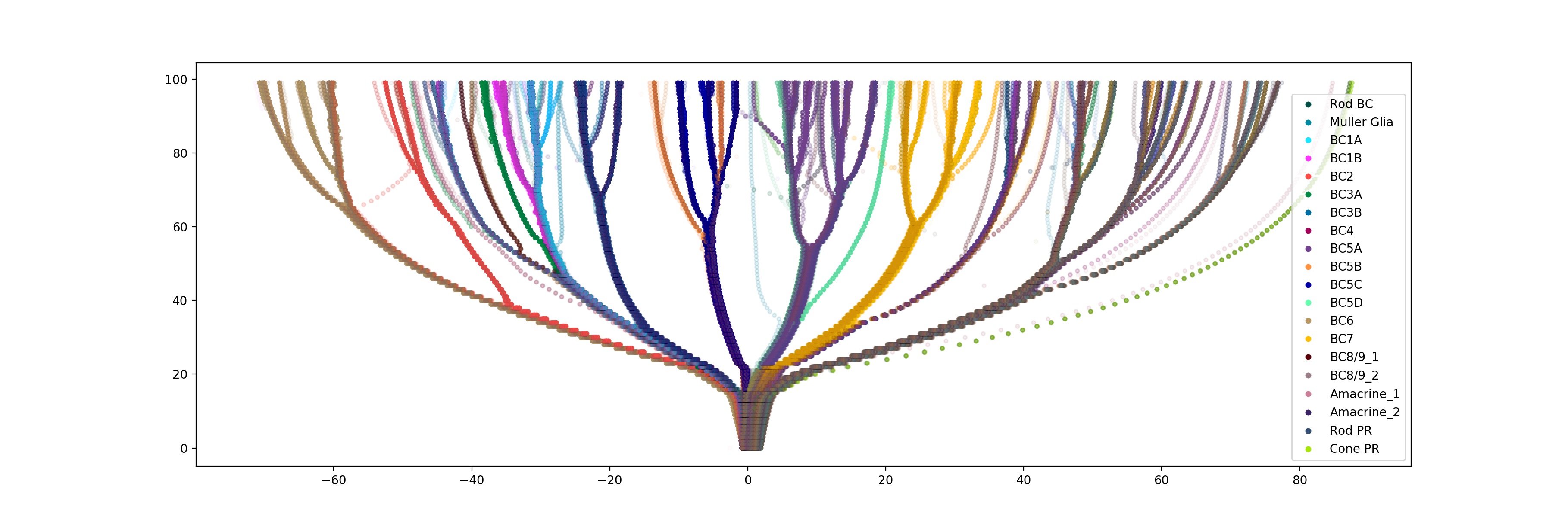}
    \caption{scRNA-seq tree-SNE embedding colored by the labels from Shekhar et al. (2016).}
    \label{fig:scRNA-seq-tree-SNE}
\end{figure}

\subsubsection{Swiss roll}

\begin{figure}[htp]
    \centering
    \captionsetup{width=.9\linewidth}
    \begin{minipage}[b]{0.4\textwidth}
    \includegraphics[width=200pt]{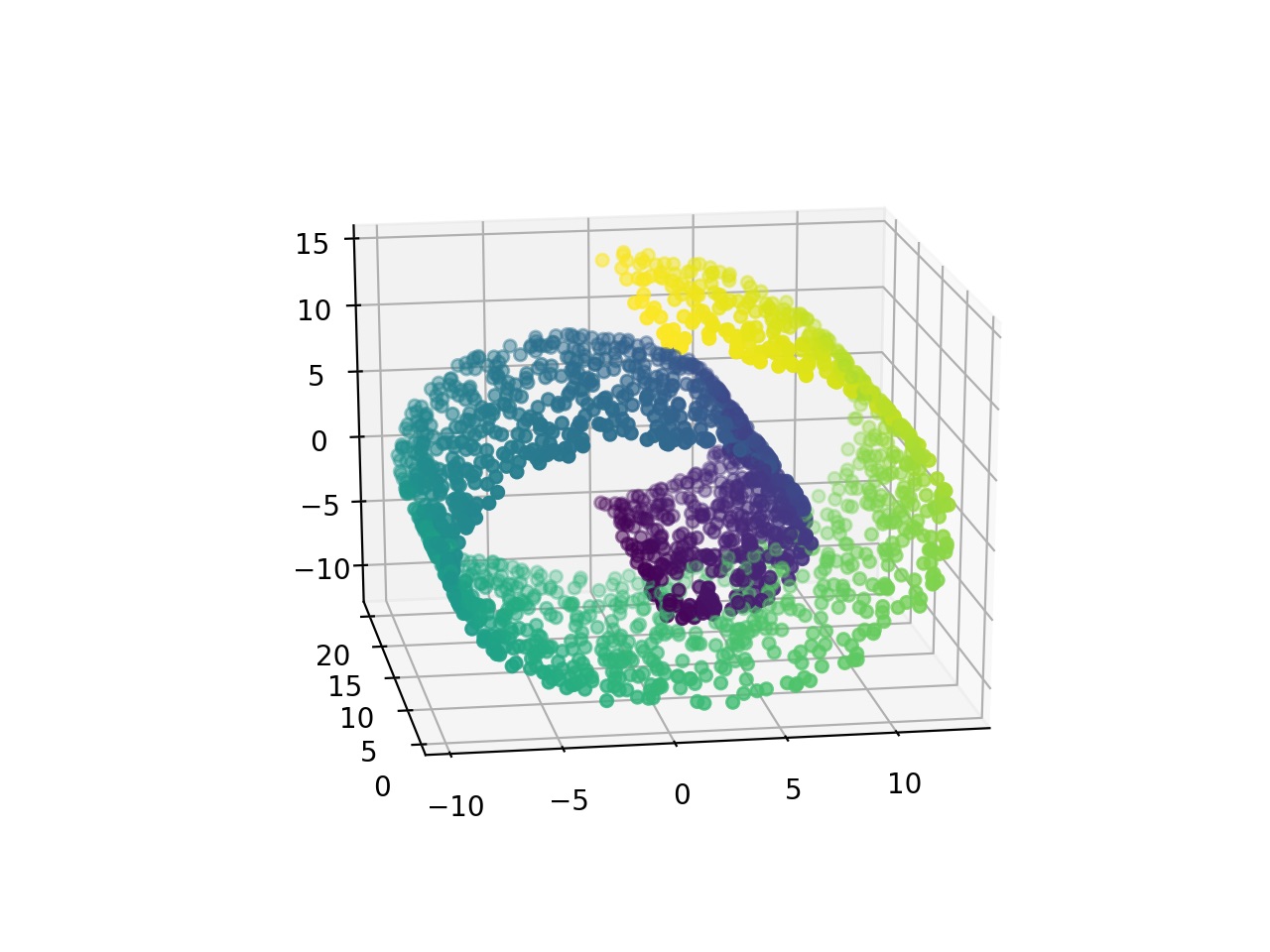}
    \end{minipage}
    \begin{minipage}[b]{0.4\textwidth}
    \includegraphics[width=200pt]{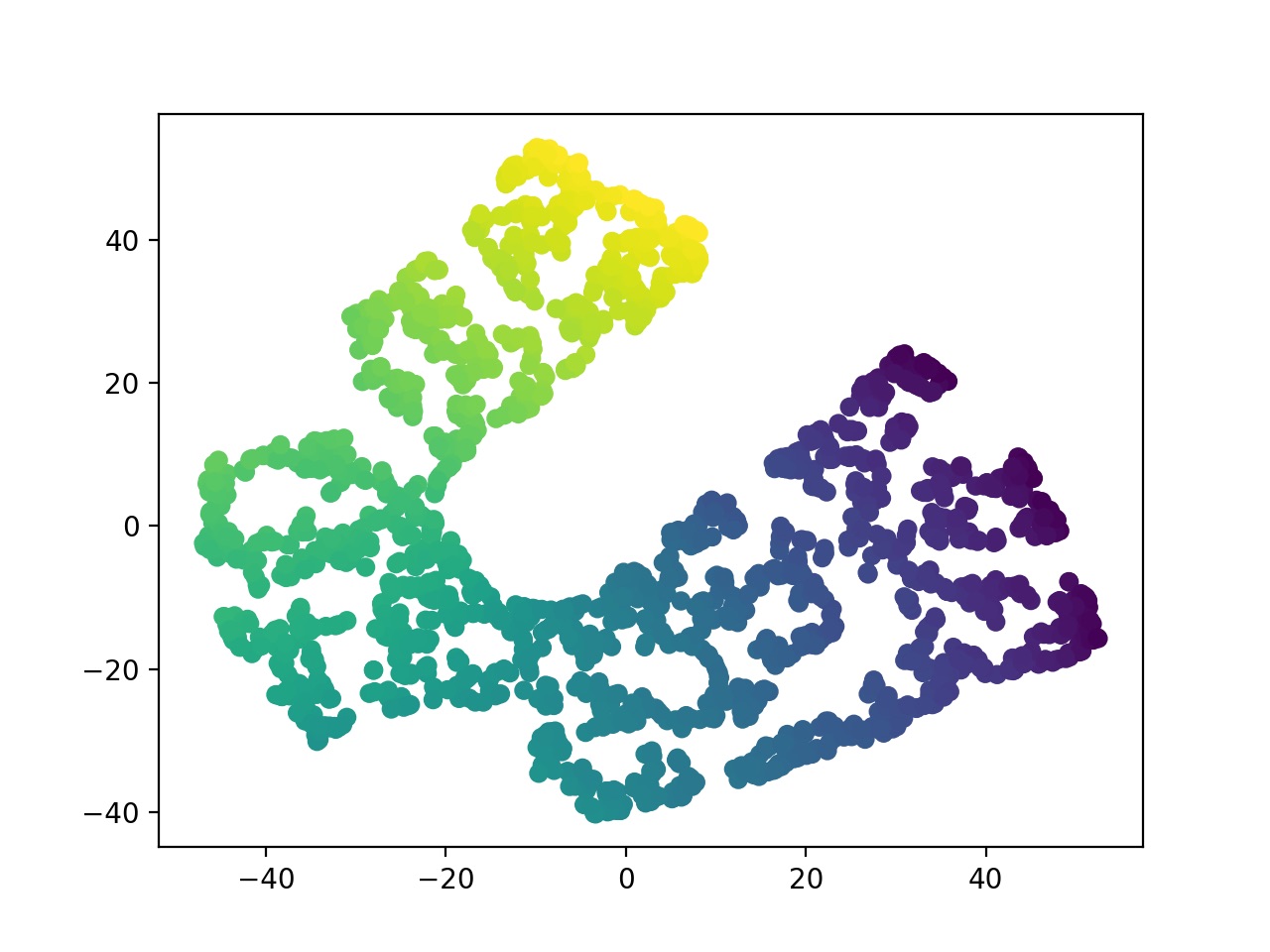}
    \end{minipage}
    \caption{Swiss roll data set in full 3-D (left) and its $t$-SNE embedding (right), both colored by labels corresponding to the position of points along the major axis of the manifold.}
    \label{fig:swiss-roll-t-SNE}
\end{figure}

\begin{figure}[H]
    \centering
    \captionsetup{width=.9\linewidth}
    \includegraphics[width=500pt]{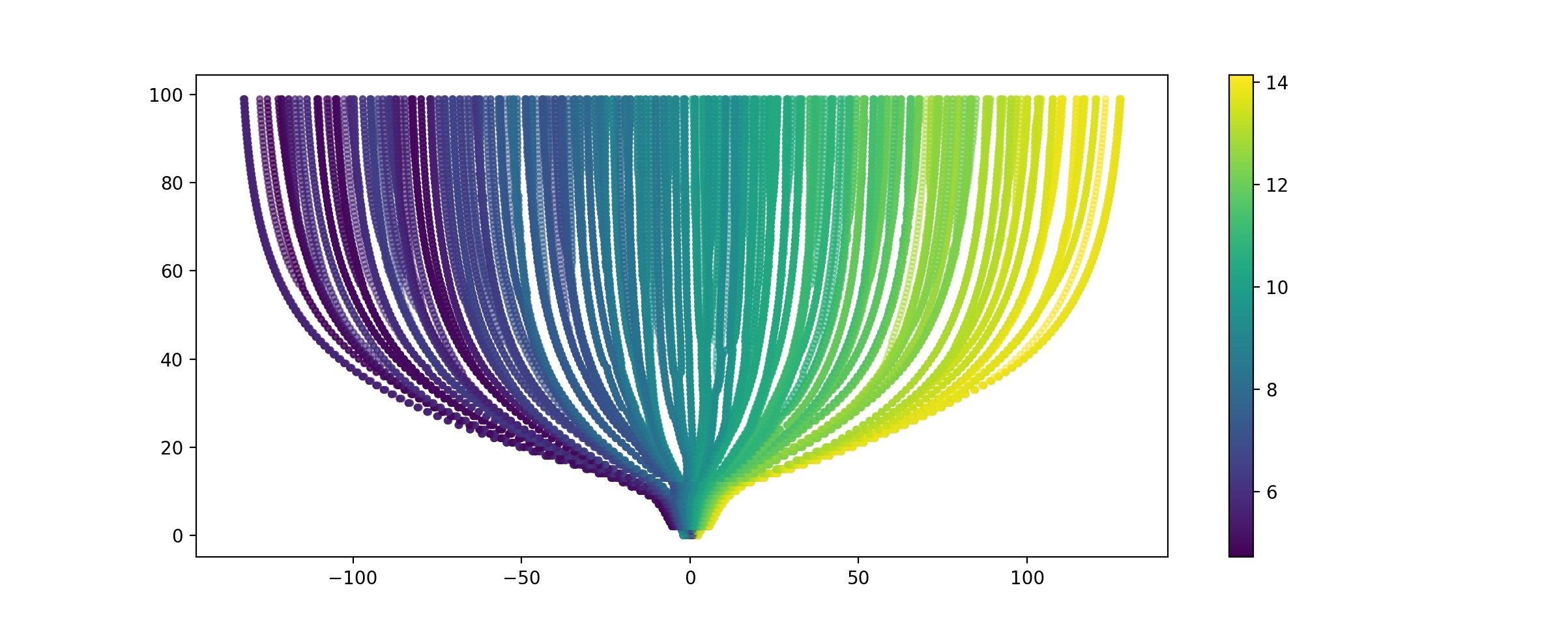}
    \includegraphics[width=200pt]{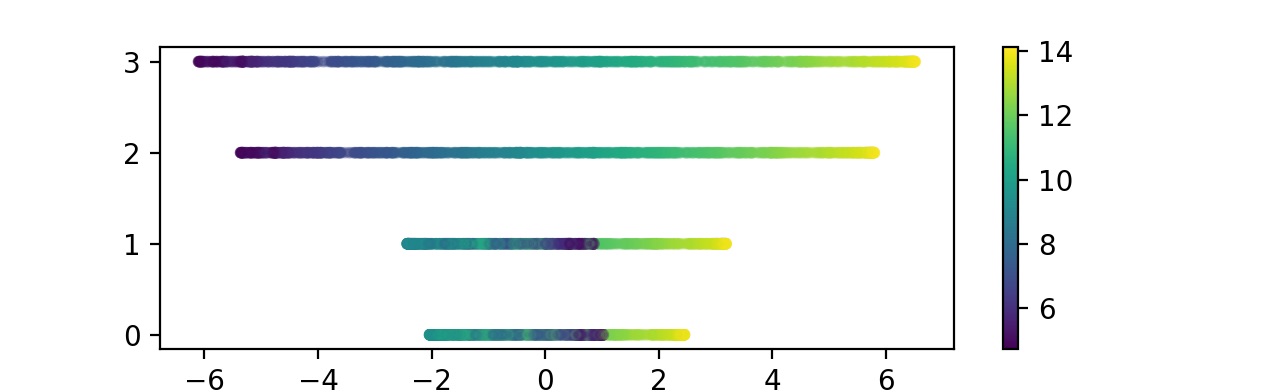}
    \caption{Swiss roll tree-SNE embedding with 100 layers (top) and just the bottom four layers (bottom), again colored by labels corresponding to the position of points along the major axis of the manifold.}
    \label{fig:swiss-roll-tree-SNE}
\end{figure}

A common concern about $t$-SNE is that it does not fully capture some standard artificial data sets, including the famous Swiss roll (Linderman and Steinerberger 2019). In the case of the Swiss roll, $t$-SNE does not fully unwind the manifold, as seen in Figure \ref{fig:swiss-roll-t-SNE}. However, tree-SNE does for the most part unwind the Swiss roll, as seen in the top plot of Figure \ref{fig:swiss-roll-tree-SNE}. Notice how on the bottom two levels of the embedding, the order of the points is wrong in the same way that it is wrong in the $t$-SNE embedding, with purple in the middle as opposed to at the end. However, tree-SNE's multi-scale approach fixes this issue and places the purple segment where it should be, correctly capturing the manifold. This phenomenon is highlighted in the bottom plot of Figure \ref{fig:swiss-roll-tree-SNE}, which shows only the first four levels of the tree-SNE plot.

Additionally, notice that there is very little cluster separation in the tree-SNE embedding of the Swiss roll in Figure \ref{fig:swiss-roll-tree-SNE}, and no well-defined branching. This shows that tree-SNE does not artificially introduce clusters or hierarchical structures when applied to data lacking this type of organization. We illustrate this point further in the next section (Section \ref {sec:noise}). This suggests that the clusters that emerge in other datasets are unlikely to be artifacts of noise, as otherwise similar clusters would have emerged here.

Although we demonstrate tree-SNE on the Swiss roll data set because it is a common "toy" data set and the behavior of tree-SNE when applied to the Swiss roll is interesting, tree-SNE is by no means a dimensionality reduction method. It is designed to visualize hierarchical cluster structures and is limited in its ability to represent manifolds.  

\subsection{Random noise} \label{sec:noise}

\begin{figure}[htp]
    \centering
    \captionsetup{width=.9\linewidth}
    \includegraphics[width=500pt]{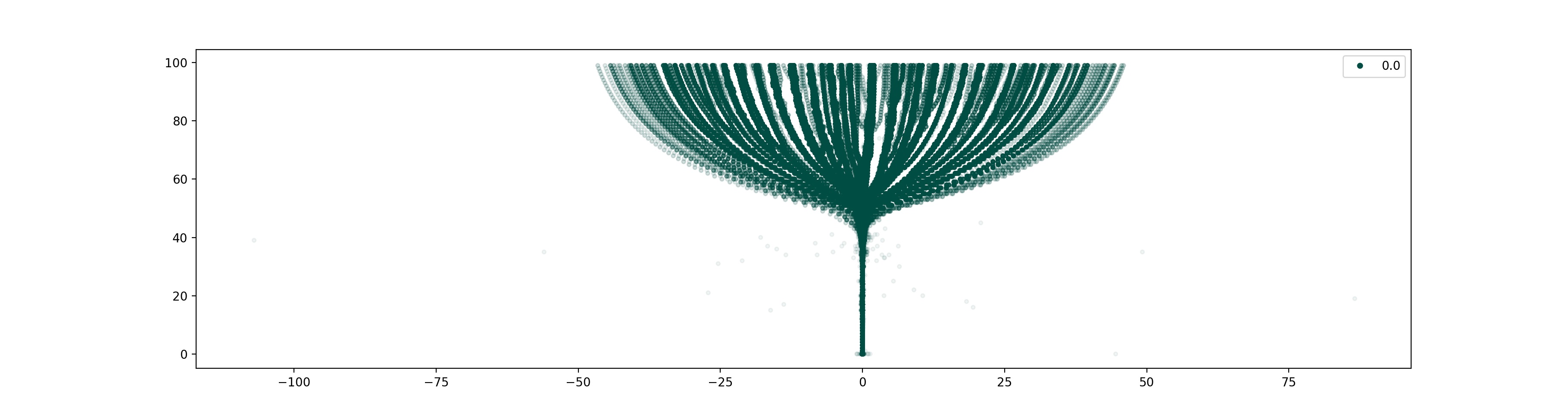}
    \caption{Uniform noise tree-SNE embedding.}
    \label{fig:noise-tree-SNE}
\end{figure}

To demonstrate that tree-SNE does not artificially introduce clusters or hierarchical structure into data lacking those structures, we run tree-SNE on 5,000 random samples drawn from a 100-dimensional uniform distribution over the range [0,1). The results are displayed in Figure \ref{fig:noise-tree-SNE}. Observe how, unlike the tree-SNE visualizations on meaningfully organized data, the tree does not begin to branch until level 50, and the subdivisions thereafter are long, without further branching. This shows that tree-SNE does not tend to produce clear cluster separation or hierarchical structure by mere artifact. Further note that there are many points seemingly randomly scattered in the embedding, which does not tend to appear in well-organized data.

\begin{figure}[htp]
    \centering
    \captionsetup{width=.9\linewidth}
    \includegraphics[width=\textwidth]{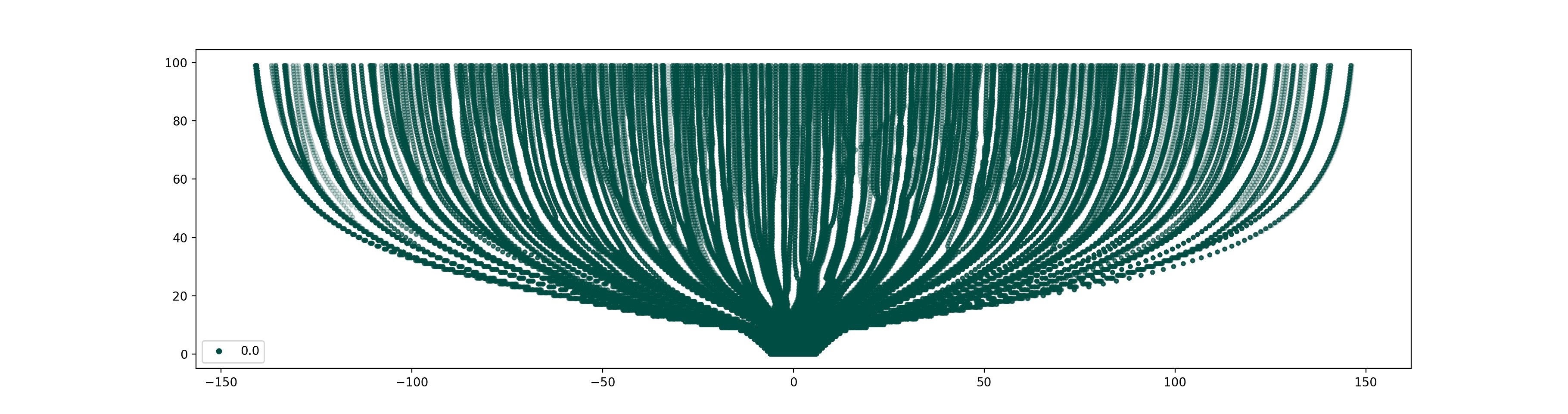}
    \includegraphics[width=200pt]{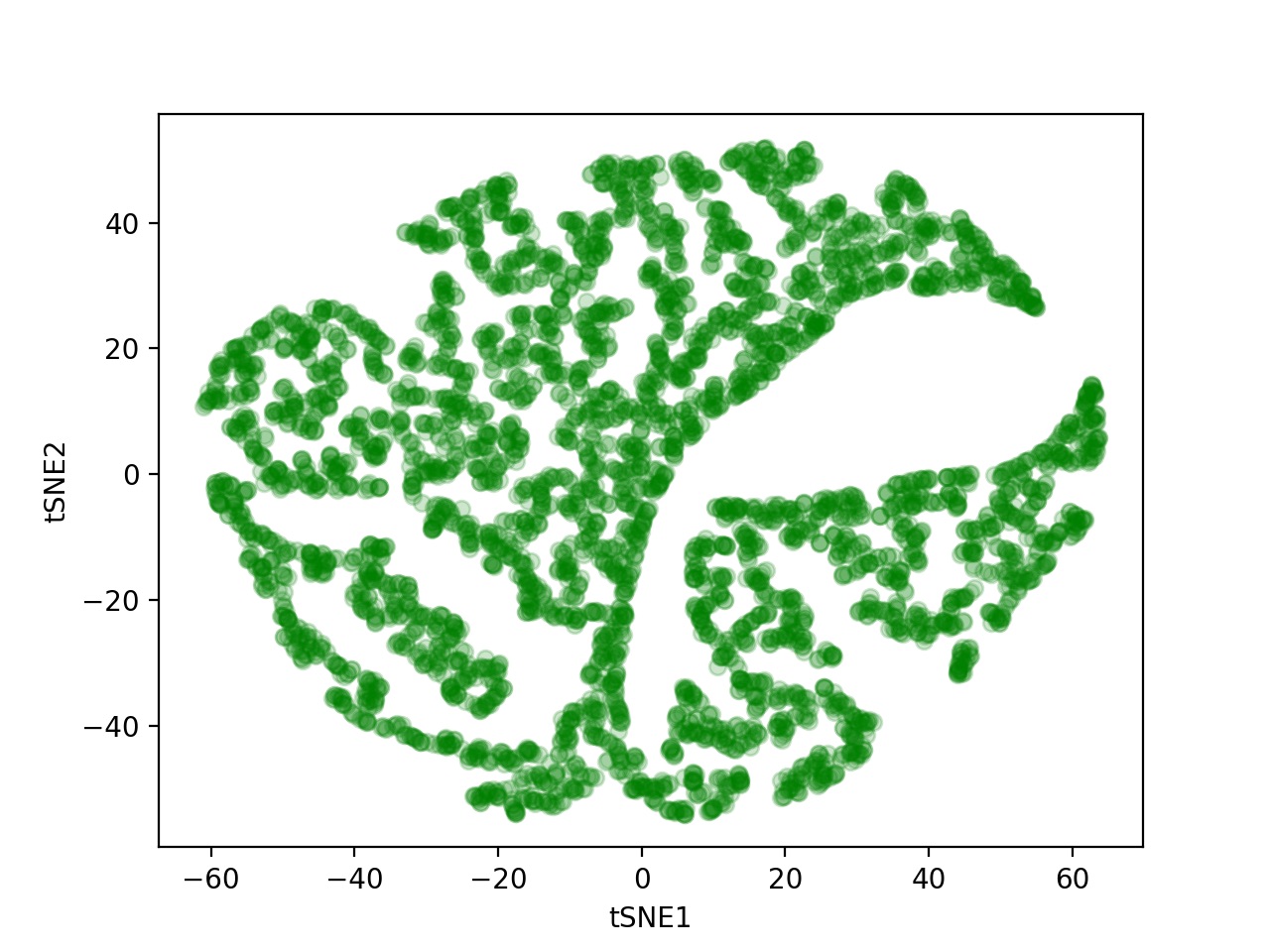}
    \caption{Random samples from a plane in 100-dimensional space visualized by tree-SNE (top) and $t$-SNE (bottom).}
    \label{fig:plane-noise}
\end{figure}

In Figure \ref{fig:plane-noise}, we generate 5,000 samples of uniform noise in a two-dimensional plane in 100-dimensional space, then embed the samples with both tree-SNE and $t$-SNE. The tree-SNE plot does not have as long of an initial stem as it does in Figure \ref{fig:noise-tree-SNE}, because the data lies along a plane rather than completely randomly throughout the 100 dimensions. The tree-SNE plot still has relatively uniformly distributed branches without a clear hierarchy, which does not tend to happen with organized data. The standard two-dimensional $t$-SNE embedding is provided for comparison. The $t$-SNE embedding gives the impression of clumping or clustering, despite the fact that the data is completely random, whereas tree-SNE does not show any clusters. 

Together, these experiments on random noise illustrate that tree-SNE does not introduce artificial structure into unstructured data, and the tree-SNE plots of noise have a characteristic profile that is distinguishable from the plots of well-organized data. 

\section{Conclusion}

We present tree-SNE, a novel hierarchical visualization method for high-dimensional data, and a companion clustering method, alpha-clustering. Tree-SNE provides a way to explore the organization of data at multiple levels of granularity, which is not available in many other visualization techniques. Alpha-clustering is built on top of the tree-SNE embedding and provides quantitative validation that tree-SNE reveals meaningful structures in the data. It also achieves NMI scores on various data sets that are competitive with state-of-the-art clustering approaches, without ever needing to specify the number of clusters. Finally, we demonstrate the viability of this approach on several diverse real-world data sets. Our analysis suggests that tree-SNE offers promise in terms of yielding insights not currently revealed by other visualization approaches. 

In the future, we intend to modify the tree-SNE algorithm to be able to leverage neural network embeddings, with the hope that this would help us be more competitive with neural network-based approaches, particularly on image or sequential data. We hope to create a means for out-of-sample extension of alpha-clustering such that the clustering approach can be used for more than exploratory data analysis, perhaps leveraging previous work on parametric $t$-SNE (van der Maaten 2009). We also want to explore the idea of a variation of alpha-clustering that varies scale ($\alpha$) differently for different branches, acknowledging that a global scale may not be optimal for all data sets. In addition, we want to investigate using alternative clustering approaches instead of subsampled SNN-based spectral clustering, perhaps leveraging the Nyström method. And finally, we want to extend tree-SNE to support stacked two-dimensional $t$-SNE plots.

\section{Acknowledgements}

The authors thank Stefan Steinerberger for inspiration, support, and advice; George Linderman for enabling one-dimensional $t$-SNE with degrees of freedom < 1 in the FIt-SNE package; Scott Gigante for data pre-processing and helpful discussions of visualizations and alpha-clustering; Smita Krishnaswamy for encouragement and feedback; and Ariel Jaffe for discussing the Nyström method and its relationship to subsampled spectral clustering.

\section{Appendix}

\subsection{Appendix A: Methods for subsetting scRNA-seq data with alpha-clustering} \label{sec:shekhar-appendix}

\begin{figure}[htp]
    \centering
    \captionsetup{width=.9\linewidth}
    \includegraphics[width=\textwidth]{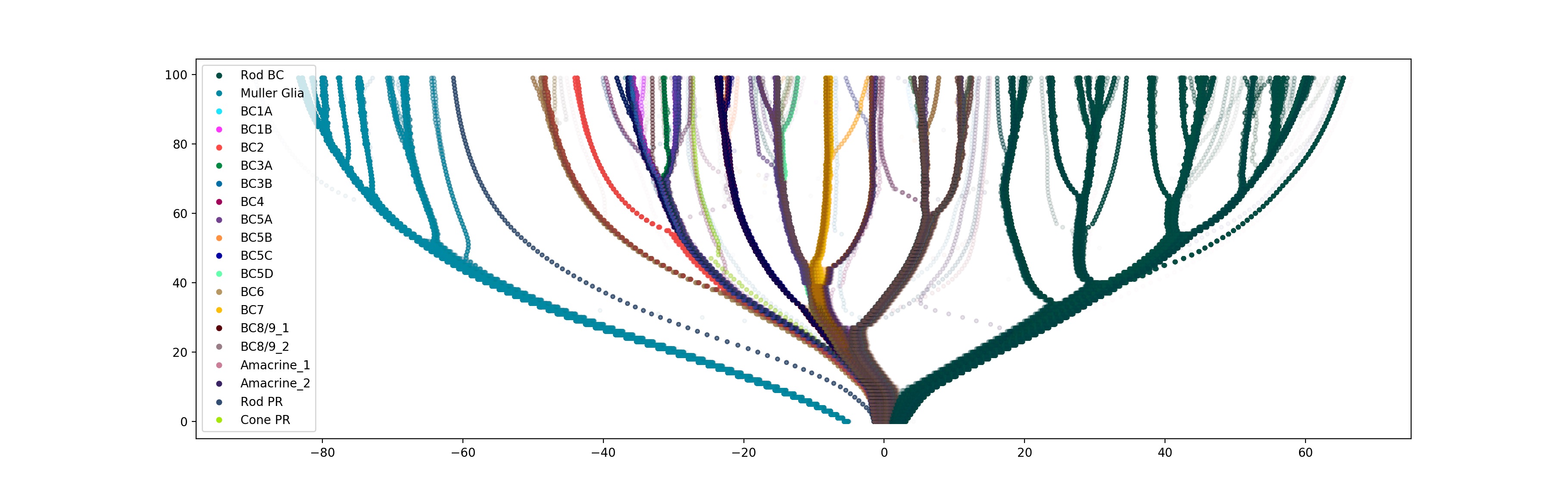}
    \caption{Tree-SNE embedding of the whole scRNA-seq data set, colored by labels from Shekhar et al (2016).}
    \label{fig:RNA-full-tree-SNE}
\end{figure}

In Section \ref{sec:shekhar}, we applied tree-SNE to a subset of a scRNA-seq data set from retinal cells (Shekhar et al. 2016). Now, we demonstrate using alpha-clustering to partition the full data set from Shekhar et al. to obtain the subset of interest used in Section \ref{sec:shekhar}. Prior to running tree-SNE, we use PCA to reduce the dimensions of the data set from 24,576 features to 100 dimensions to increase the efficiency of the algorithm. We start with the tree-SNE embedding of the full scRNA-seq data set, in Figure \ref{fig:RNA-full-tree-SNE}, colored by the labels from Shekhar et al. Notice that the Muller Glia (teal, leftmost branch) and Rod BC (dark green, rightmost branch) cell types take up a lot of space in the visualization, crowding the main group of cone bipolar cell (BC) subtypes of interest. We would like to remove the Müller glia and rod BCs to reduce this crowding, and we will use alpha-clustering labels to partition the data to do so. 

\begin{figure}[htp]
    \centering
    \captionsetup{width=.9\linewidth}
    \includegraphics[width=\textwidth]{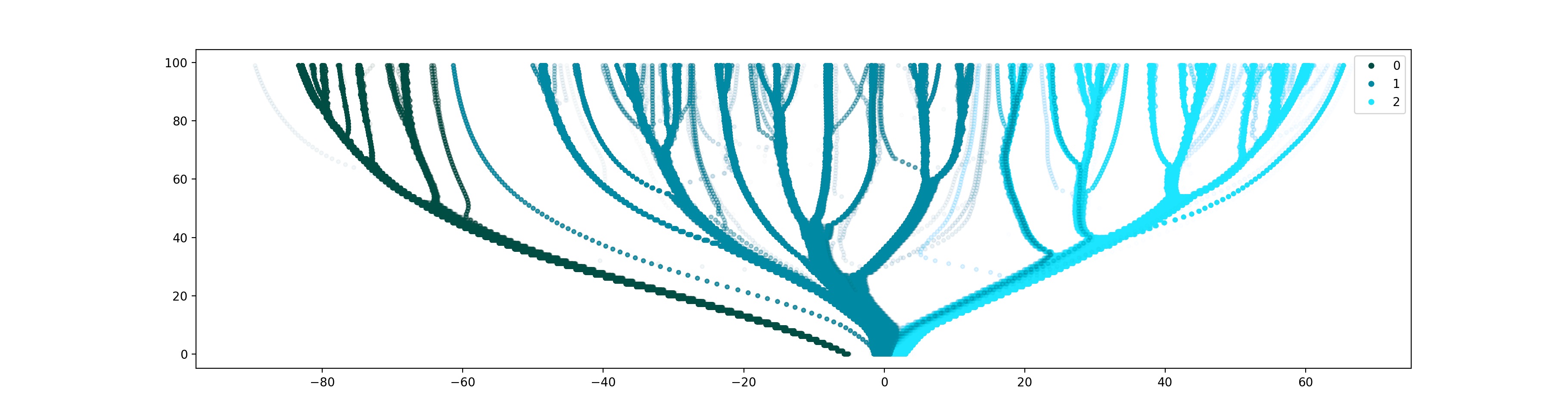}
    \caption{Tree-SNE embedding of the whole data set of scRNA-seq data, colored by alpha-clustering labels.}
    \label{fig:RNA-clusters-tree-SNE}
\end{figure}

Figure \ref{fig:RNA-clusters-tree-SNE} shows the same tree-SNE embedding of the full scRNA-seq data set, this time colored by alpha-clustering labels. Alpha-clustering automatically determined that the optimal number of clusters was three, and these three clusters correspond to (left to right) Müller glia, cone BC subtypes and photoreceptor (PR) cells, and rod BCs. 

\begin{figure}[htp]
    \centering
    \captionsetup{width=.9\linewidth}
    \includegraphics[width=\textwidth]{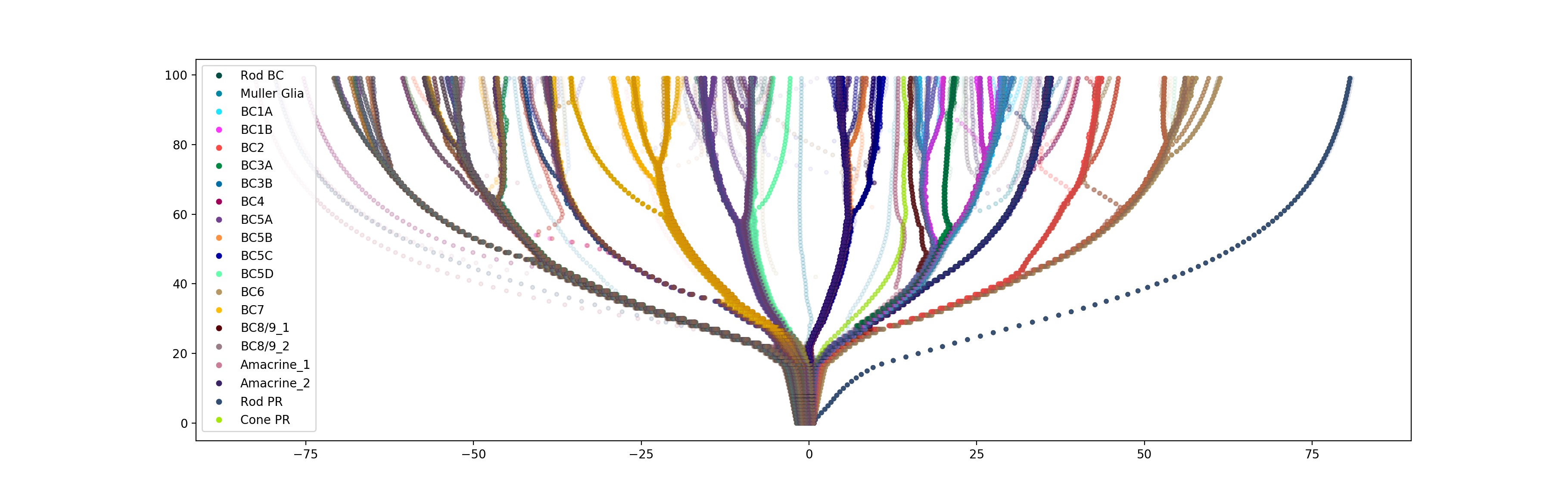}
    \caption{Tree-SNE embedding of a subset of the scRNA-seq data, excluding cells that are not of interest.}
    \label{fig:RNA-subset-tree-SNE}
\end{figure}

Since alpha-clustering has already partitioned the data, it is simple to remove clusters 0 and 2 and re-apply tree-SNE to only the observations within cluster 1, allowing for a visualization that has more space to preserve the hierarchical organization of the cells of interest. This new embedding is shown in Figure \ref{fig:RNA-subset-tree-SNE}. Notice that on the far right, there is still a small group of rod PR cells present in this plot, which are not a cell type of interest. 

\begin{figure}[htp]
    \centering
    \captionsetup{width=.9\linewidth}
    \includegraphics[width=\textwidth]{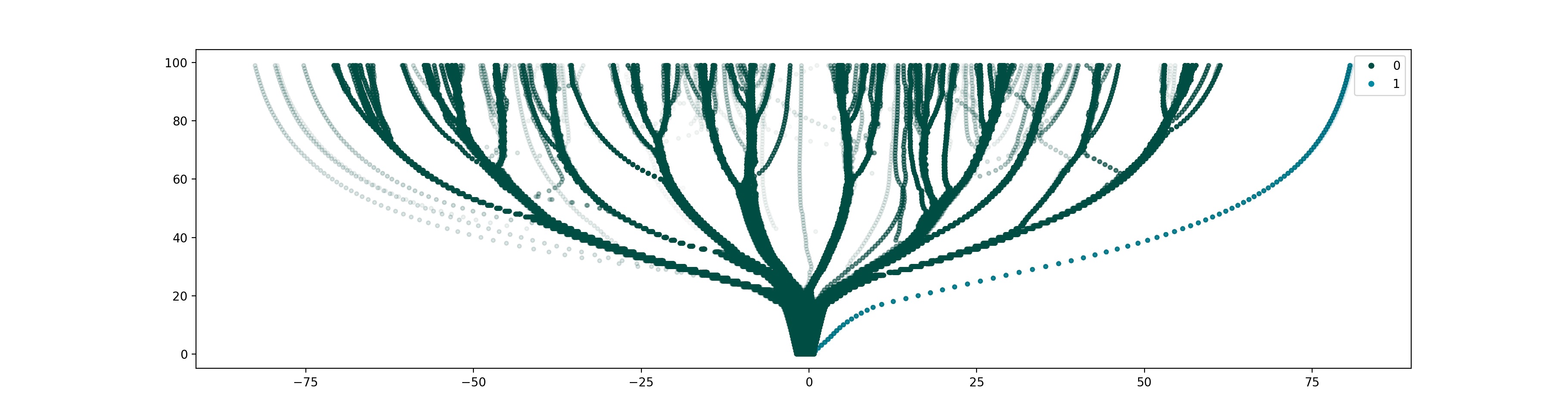}
    \caption{Tree-SNE embedding of a subset of the scRNA-seq data, colored by alpha-clustering labels.}
    \label{fig:RNA-subset-clusters-tree-SNE}
\end{figure}

Running alpha-clustering on the tree-SNE embedding from Figure \ref{fig:RNA-subset-tree-SNE} yields an optimal cluster assignment of just two clusters; again, the number of clusters was automatically detected by the algorithm. Figure \ref{fig:RNA-subset-clusters-tree-SNE} is colored by the alpha-clustering labels, showing that the two clusters discovered from this subset of the data correspond to the majority of the tree (left) and the rod PR cells (right). 

\begin{figure}[H]
    \centering
    \captionsetup{width=.9\linewidth}
    \includegraphics[width=\textwidth]{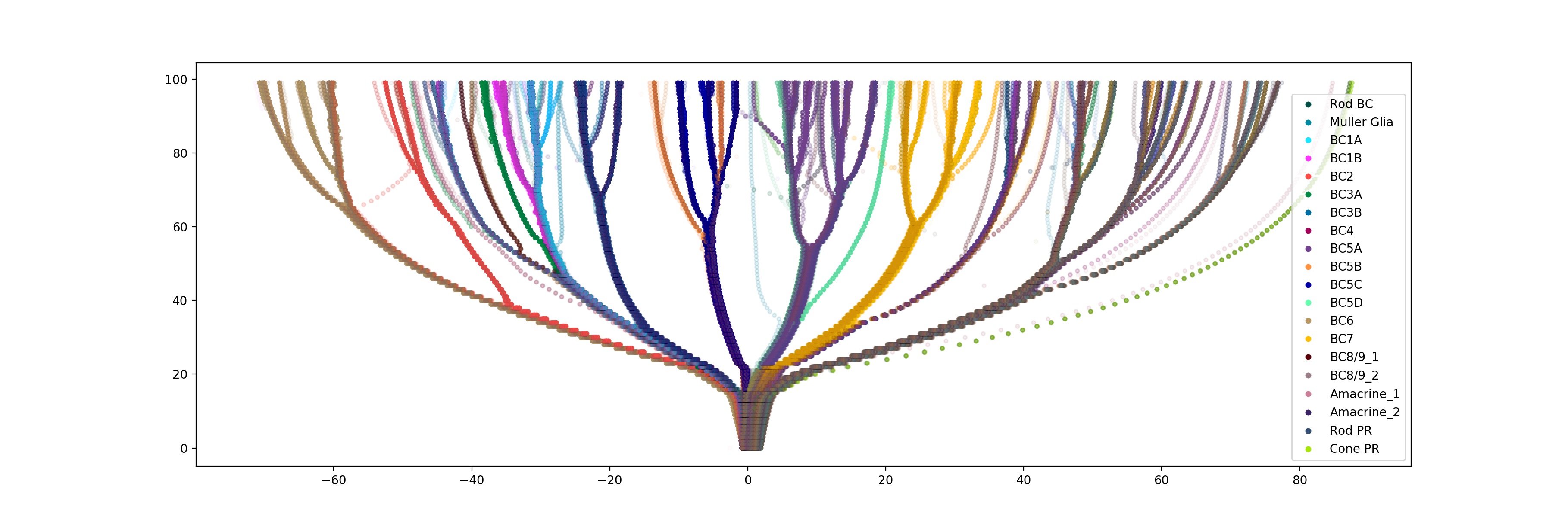}
    \caption{Tree-SNE embedding of only the cone BC subtypes in the scRNA-seq data (with a few rogue cone photoreceptor cells on the right side), colored by labels from Shekhar et al. (2016).}
    \label{fig:RNA-subsubset-tree-SNE}
\end{figure}

The alpha-clustering labels make it trivial to remove cluster 1 and re-embed the remaining data from cluster 0. For this final round of tree-SNE, we also reduce the number of PCA components used to 37 to more closely match the methods from Shekhar et al. (2016) and enable more direct comparison. The final tree-SNE embedding is shown in Figure \ref{fig:RNA-subsubset-tree-SNE} (which is the same as Figure \ref{fig:scRNA-seq-tree-SNE} shown previously), and is discussed in more detail in Section \ref{sec:shekhar}.

The ability to automatically detect, partition, and remove particular significantly different subsets of the data is, to the best of our knowledge, unique to tree-SNE and alpha-clustering. This allows a researcher to very easily select and hone in on subsets of interest in their data.

\bibliographystyle{unsrt}  


\end{document}